\documentclass{article}
\usepackage[table,xcdraw]{xcolor}
\usepackage{natbib}
\usepackage{comment}
\usepackage{soul}
\usepackage{rotating}
\usepackage[utf8]{inputenc} 
\usepackage[T1]{fontenc}    
\usepackage{hyperref}       
\hypersetup{
    colorlinks=true,
    citecolor=blue,
    filecolor=magenta,      
    urlcolor=cyan,
    pdftitle={Overleaf Example}, 
    pdfpagemode=FullScreen,
    }
\usepackage{float}
\usepackage{url}            
\usepackage{booktabs}       
\usepackage{amsfonts}       
\usepackage{nicefrac}       
\usepackage{microtype}      
\usepackage{graphicx}
\usepackage{bm}
\usepackage{adjustbox}
\usepackage{geometry}
\usepackage{caption}
\usepackage{subcaption}
\usepackage{multirow}
\usepackage{tabu}
\usepackage{siunitx} 
\usepackage{amsmath}
\usepackage{hyperref}
\usepackage{algorithm}
\usepackage{algpseudocode}
\usepackage{diagbox}
\newcommand\mytab[1]{\begin{tabular}{@{}c@{}}#1\end{tabular}}

\makeatletter
\title{Delay Neural Networks (DeNN) for exploiting  temporal information in event-based datasets}
\author{
   Alban Gattepaille \\
   Université Côte d'Azur - CNRS - I3S \\
   \texttt{alban.colas-gattepaille@univ-cotedazur.fr} \\
   \\
   Alexandre Muzy \\
   Université Côte d'Azur - CNRS - I3S \\
   \texttt{alexandre.muzy@cnrs.fr}
}

\begin{document}
\maketitle
\begin{abstract}
In Deep Neural Networks (DNN) and Spiking Neural Networks (SNN), the information of a neuron is computed based on the sum of the amplitudes (weights) of the electrical potentials received in input from other neurons. We propose here a new class of neural networks, namely Delay Neural Networks (DeNN), where the information of a neuron is computed based on the sum of its input synaptic delays and on the spike times of the electrical potentials received from other neurons. This way, DeNN are designed to explicitly use exact continuous temporal information of spikes in both forward and backward passes, without approximation. (Deep) DeNN are applied here to images and event-based (audio and visual) data sets. Good performances are obtained, especially for datasets where temporal information is important, with much less parameters and less energy than other models.
\end{abstract}
\section{Introduction}
Deep Neural Networks (DNN) have gained more and more in complexity, power and performance to solve highly complex tasks \citep{reviewDeepCNN}. These networks abstract the functioning of biological neurons. Electrical information is integrated, computed and passed from the preceding layer to the next. As these networks can use a lot of energy \citep{energyConsumption}, and aren't biologically plausible, a new class of neural networks has emerged, Spiking Neural Networks (SNN), which tend to reproduce the spiking behavior of biological neurons.

In an SNN, each neuron is represented by an electrical membrane potential, which evolves according to incoming spikes. Once the membrane potential reaches a threshold, the neuron emits a spike and its membrane potential is usually reset. This all-or-nothing behavior reduces the number of computations because neurons possibly do not fire and thus do not activate downstream neurons. The thresholding of membrane potentials induces a discontinuity in the model, which impedes mathematical analysis and the computation of the gradient in the backpropagation algorithm and thus complicates the learning. 

Time dimension can be used in different manners in SNN, depending on the model. In \citep{kim2021visual}, using rate-coded SNN, it is possible to show that short inter-spike intervals carry information. In \citep{thorpe2001spike}, in order to better account for the precise firing times of the neurons, a new coding method has emerged, namely Time-To-First-Spike (TTFS) coding. \citet{thorpe2001spike} argue that the biological brain could make use of precise timing of the spikes, or of the order of arrival of each incoming spike. In TTFS coding, neurons are usually forced to spike only once, thus constraining the network to arrange spikes in time. Recently, the learning of synaptic delays has came along the will of using the time dimension \citep{zhang2020temporal, hazan2022memory, hammouamri2023delays, sun2023axonal}. Indeed, it can be shown that, for some type of data, delays but not weights are necessary to solve temporal logic problems \citep{habashy2024}. Moreover, temporal plasticity can be used to treat temporal information \citep{wang2024noise}. For more theoretical analyses on the interest of synaptic delays in SNN, one can refer to \citep{maass1997time, maass1997complexity, thorpe2001spike}.

We present here Delay Neural Networks (DeNN), which can be considered to be a temporal version of DNN, or an abstract SNN. As classic DNN and SNN treat electrical amplitudes (or models of), DeNN treat timing information (or a model of).  In DeNN, learning happens through synaptic delays, and an important connection between any two neurons is represented by a short delay. The firing time of a neuron is computed by directly considering the impact of each presynaptic spike onto the firing time of the postsynaptic neuron instead of thresholding a membrane potential. This allows side stepping the challenge of non-differentiability faced by SNN, and using exact temporal information from input synapses in the forward and backward passes. We show results on event-based datasets for classification tasks (video and audio). 

The \textbf{contributions} of this work are as follows: (i) We introduce a \textbf{new general framework for working in the temporal dimension with deep neural networks}, which can be adapted to deep neural network architectures ; (ii) In this framework, \textbf{different temporal kernels can be experimented} to learn \textbf{synaptic delays} instead of synaptic weights, with exact evaluation of the gradient in the temporal dimension ; (iii) \textbf{On benchmark datasets (images, videos and audio), DeNN obtains the same, or better performances than other models, but with much less parameters and less energy cost, with respect to other models}.

\section{Related works}
\subsection{Temporal coding in SNNs}
Learning temporal codes has a long history in computational neuroscience. We can cite, among many other approaches, Tempotron \citep{gutig2006tempotron} and ReSuMe \citep{ponulak2009resume}. Tempotron presents an interesting weight synaptic learning rule for a single neuron to learn for detecting locally synchronous spikes, known as Spike-Timing-Dependent Plasticity (STDP). ReSUME focuses on making (reservoir networks of) neurons learn to reproduce template signals (instructions) encoded in precisely timed sequences of spikes. Instead, the purpose of DeNN is to follow a modernized version of Time-To-First-Spike (TTFS) temporal code implementing delays, and adapting the idea to event-based datasets. DeNN neurons learn synaptic and firing delays for firing faster or slower according to the classification error of a temporal signal in a deep learning network architecture. While DeNN is currently more oriented to global deep learning mechanisms, computational neuroscience mechanisms will help in the future to improve and to better understand the global and local learning mechanisms of DeNNs.

\citet{mostafa2017supervised} was one of the first to introduce temporal coding in modern deep spiking neural networks. To do so, they derived an analytic formula to directly compute the firing times of non-leaky Integrate and Fire (IF) neurons which produce a single spike, \textit{i.e.}, with infinite refractory period. \citet{zhou2021temporal} extended this algorithm to more challenging benchmark datasets in computer vision. Their work was extended to the Spike Response Model (SRM) neurons by \citep{comsa2020temporal}, where an equation is solved in the complex field to find solutions for the good spike timing of the neurons. With the same technique, \citet{goltz2021fast} extended the work in \citep{mostafa2017supervised} to several cases of Leaky Integrate and Fire neurons (LIF). All these models have in common a methodology to solve an equation to find the timing of the first spike from a neuron (whether it is an IF neuron, a LIF, an SRM). With the same methodology, \citet{park2020t2fsnn} presented equations somewhat simpler than in \citep{goltz2021fast, comsa2020temporal}. 

Another solution is to adapt the backpropagation algorithms to event-based data \citep{zhu2022eventbackprop}. Recent works \citep{wunderlich2021gradient, forwardProp} show that an exact computation of the gradient for these models is possible, and that rate coding and temporal coding can be related through loss functions \citep{zhu2023loss}. Most of these approaches are either restricted to specific models \citep{mostafa2017supervised, comsa2020temporal, goltz2021fast} or to approximate gradients \citep{zhu2022eventbackprop}. Among approximate gradient techniques, the surrogate gradient allows SNNs to exhibit promising results \citep{esser2016surrogate, bellec2018surrogate,  neftci2019surrogate, tavanaei2018review}. However, this technique remains an approximation of the thresholding function used in the forward pass by continuous functions in the backward pass. 

Another possibility that has been developed is to map classic neurons to spiking neurons, as in \citep{rueckauer2018conversion, park2021training, kheradpisheh2020temporal}. These works allow simpler backpropagation algorithms, since they can leverage better the properties of analog networks. They are however still restricted to one model of neuron and can hardly be generalized to other models. 
\subsection{Synaptic delays}
\noindent To achieve temporal coding, Single spike SNNs \citep{mostafa2017supervised, zhang2019tdsnn, zhou2021temporal, comsa2020temporal, goltz2021fast, park2020t2fsnn} intuitively seem also to be a good framework to learn delays between neurons. Modelling the delays into neural networks can be tracked back at least to 1989 with Time-Delay Neural Networks (TDNN) \citep{hintonTDNN}. In this work, connections have several synaptic terminals, each with its own fixed delay and variable weight thus leading to an exploding number of parameters. \citet{bohte2002error} derived an approach called Spike Prop, with the same architecture for synaptic terminals and the same drawbacks in terms of memory and computations. 

Simpler delay-based models have then been developed. In \citep{schrauwen2004extending, belatreche2006constraints, shrestha2018slayer, hammouamri2023delays, sun2023axonal} one trainable delay was implemented for each synaptic connection, alongside with synaptic weights. In \citep{taherkhani2015edl, taherkhani2015dl}, single output neurons are trained to fire a spike train at desired times. Also, \citet{taherkhani2015dl} only allow delays to be increased, which seems biologically implausible. \citet{zhang2020temporal} presented an interesting joint synaptic delay-weight plasticity algorithm, and confronted it to a real-world dataset for speech recognition. More recent works tend to get rid to the single spike constraint, as it is not reliable for event based datasets \citep{yu2023multispike, hammouamri2023delays, sun2023axonal, grappolini2023delays, deckers2024delays, wang2024delays}.

In most of these delay-based works, every synapse has two parameters: a weight and a delay, effectively doubling the memory cost with respect to analog networks. To the best of our knowledge, \citep{hazan2022memory} is the only work presenting a weightless spiking neural network, where learning happens only through synaptic delays. The authors used a learning rule derived from Spike-Timing-Dependent Plasticity (STDP), and confronted their network to a classic image classification task. However, their network exhibits low accuracy and memory performances. 
\section{Methods}
\label{sec:math}
\subsection{Forward pass}
\label{sec:forward}
In our network, instead of weights $w_{ij}$, we use positive delays which represent a simple time delay between two presynaptic and postsynaptic neurons. In DNN, if a connection between two neurons is important  (inversely not important), its weight $|w_{ij}|$ is high (inversely low), while in DeNN this corresponds to a small (inversely high) delay $|d_{ij}|$. In order to ensure that the delays $d_{ij}$ stay positive during the back-propagation steps, we effectively declare signed-delays $d^{s}_{ij} \in \mathbb{R}$, and then compute the delay as a simple Gaussian of the signed-delays: 
\begin{equation}
    \label{eq:delay}
    d_{ij} = \exp \Big( - \Big(\frac{d_{ij}^{s}}{\sigma_{j}} \Big)^{2} \Big) 
\end{equation}
where $\sigma_{j}$ is a parameter learnt for each neuron. 
\textbf{DeNN's neurons generic definition simply consists of Equation \ref{eq:sum_gen}}. The firing time of postsynaptic neuron $j$ is computed as a function of the spiking times of the preceding layer ($t_i > 0$) and synaptic delays:
\begin{equation}\label{eq:sum_gen}
    t_j = \sum_i f(t_i, d_{ij}^{s})
\end{equation}
The function $f$ represents the synaptic impact of presynaptic neuron $i$ onto the spiking time of postsynaptic neuron $j$, $t_j$.  
Essentially, for any synaptic input received, which is excitatory (resp. inhibitory), every neuron $j$ fires earlier (resp. later). In DeNN, low (inversely high) spiking times would correspond to high (inversely low) activations in DNN.

\textbf{Many DeNN's neuron functions $f$ can be derived from Equation \ref{eq:sum_gen}}.  
The following has been chosen after considering arrival times $t_i + d_{ij}$ for presynaptic neuron $i$ and postsynaptic neuron $j$:
\begin{equation}\label{eq:sum_kappa}
    t_{j} = \sum_i sign(d_{ij}^{s}) \Big[ \kappa(t_{i} + d_{ij}) - \kappa(t_{i} + 1) \Big]
\end{equation}
where $\kappa$ is a strictly decreasing positive function representing the impact of the incoming spike onto the firing time of the postsynaptic neuron. The term $- \kappa(t_i + 1)$ represents an incompressible delay because it decreases the activity of the synapse, and tackles a discontinuity at $d_{ij} = 1$ (see Section \ref{sec:continuity} for more details). The sign of the signed-delay, between a presynaptic neuron $i$ and a postsynaptic neuron $j$, is taken into account to represent the type of the synapse: whether it is an excitatory synapse (negative synapses, to decrease the time of activation), or an inhibitory one (positive synapse). We found that a satisfactory kernel $\kappa$ was the exponential kernel $\kappa(x) = e^{-x}$. Thus, with this kernel, each spike is exponentially more important than subsequent ones. This corresponds to a balanced mix between TTFS and rank-order coding \citep{thorpe2001spike}. In that sense, DeNN is a temporal abstraction of the behavior of biological neurons. Note that any continuous positive decreasing function can be used for kernel $\kappa$, but the exponential one happens to work well in practice.
\subsection{Standardization and temporal ReLU function}
\label{sec:std_relu}
\begin{figure}[!t]
  \centering
  \includegraphics[width=0.8\linewidth]{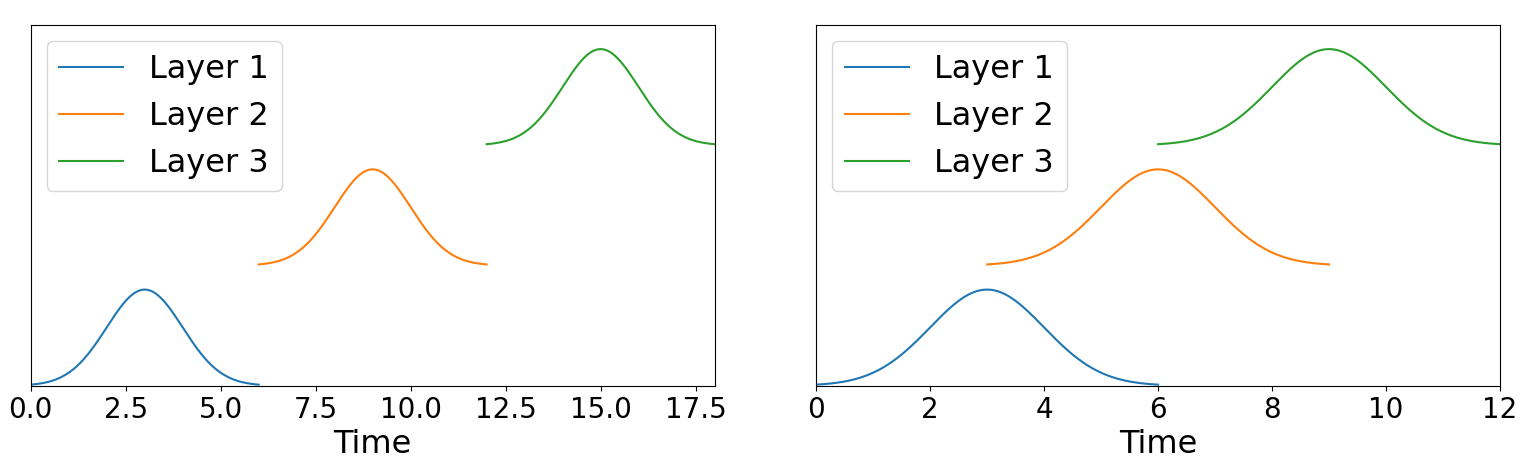}
  \vspace{-0.3cm}\caption{Slow (left) and fast (right) regimes of the DeNN. Each layer outputs spikes after an integration phase, which duration is calibrated by $q$. If $q=1$, then each layer has to wait until every neuron of the preceding layer has emitted a spike, which corresponds to the slow regime. To reduce the latency of the model, it is possible to decrease $q$ (fast regime), so that each layer can ignore the slowest neurons of the preceding layer.}
  \label{fig:regimes}
\end{figure}

It is possible then to cancel (or not) every neuron that fires after some value (typically the median of the spiking times in the layer), forcing them to an infinite time. This process is equivalent to a simple lateral inhibition, where the first neurons of a layer to spike in time impede the neighbouring slower ones. It is equivalent to sending a signal to the neurons of the layer after the spike corresponding to the $q$-th quantile to make them silent. With $q=1$, all spikes are fired, this corresponds to a slow regime (Figure \ref{fig:regimes}, left). With $q=0.5$, all spikes after the median (or the average in case of gaussian distribution, see Section \ref{fig:neurons_before_std} for an experimental illustration) are silent, and this corresponds to the fast regime (Figure \ref{fig:regimes}, right).\textbf{In software systems, the DeNN equations simply become:}
\begin{equation}
    \label{eq:software_denn}
    \begin{split}
        t_j &= \sum_i sign(d_{ij}^{s}) \Big[ \kappa(z_{i} + d_{ij}) - \kappa(z_{i} + 1) \Big] \\
        z_j &= std(t_j)
    \end{split}
\end{equation}
where $std$ is the standardization process, where we subtract the mean and divide by the standard deviation of the distribution. As shown in Section \ref{sec:standardization}, taking $\kappa(z_{i} + d_{ij})$ or $\kappa(t_{i} + d_{ij} - t_q)$ is ``almost'' equivalent, up to a division by standard deviation. For the fast regime ($q=0.5$), a temporal ReLU can be defined as:
\begin{equation}
  TempReLU(z) =
    \begin{cases}
      z    & \text{if $z<0$}\\
      +\infty & \text{otherwise}
    \end{cases}     
\end{equation}
\subsection{Events Preprocessing: event2time algorithm}
\label{sec:neuro-preproc}
\noindent When event-based datasets consist of images obtained from event-based cameras, pixel-level intensity changes are captured as events. Since the number of events increases with the temporal resolution of the camera, the number of events can get large.

\begin{figure}[!h]
  \centering
  \includegraphics[width=0.8\linewidth]{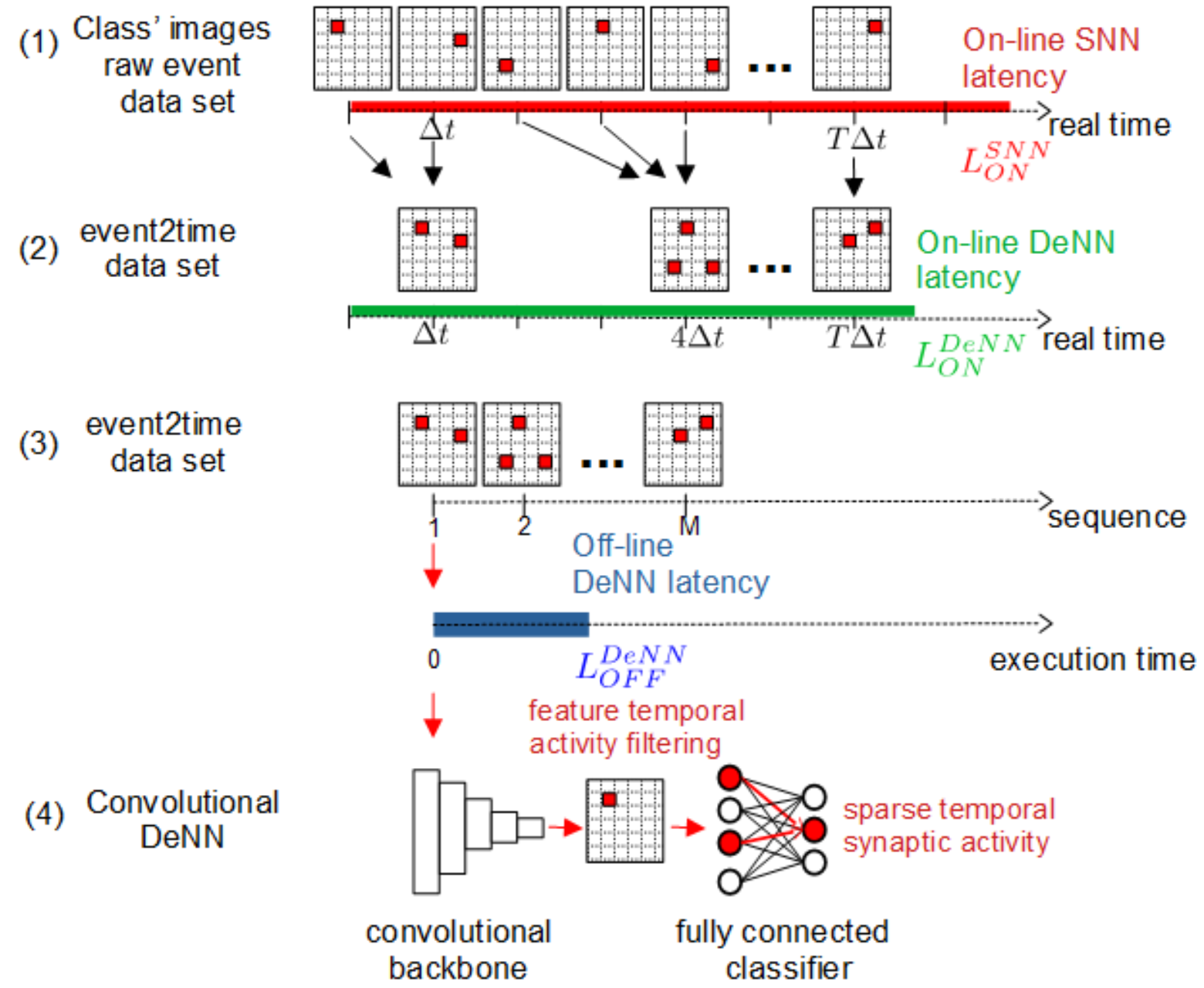}
  \caption{DeNN full pipeline: (1) Events arrive every $\Delta t$ timestep, where $\Delta t$ is the precision of the neuromorphic camera, and are used such as by SNN. (2) Using event2time algorithm, events are aggregated over $T\Delta t$ timesteps, for one sample, and fed to the network in an on-line manner. (3) After all events are computed, the sequence of $M$ images can be simulated much faster jumping from one image to the other without waiting for every $\Delta t$ timestep, giving rise to an off-line latency. (4) Feed-forward network used for each image.}
  \label{fig:story_telling}
\end{figure}

To deal efficiently with event-based datasets, an algorithm for the pre-processing of events, inspired from \citep{zeigler2004discrete}, was developed \citep{fred}. This algorithm greatly reduces the number of events in the datasets while tracking the most relevant pixels' activity. In event-based datasets, a data consists of three coordinates $(t, p, \mathbf{x})$ where $t$ is the time of the event, $p \in \{-1, +1\}$ is the polarity of the event, and $\mathbf{x}$ is the position of the pixel in the image. Our algorithm, called \textit{event2time}, accumulates events on each cell over time.  Each cell $i$ stores the timing of the events in a list $L_i$, and when more than $2rN$ ($0<r<1$, for $N$ total pixels in the image, and two polarities) cells are active (\textit{i.e.}, have stored one or more events), an array is built with:
\begin{equation}
    \label{eq:event_to_time}
    \begin{split}
    t_{i} = \frac{\max L_{i} - \min L_{i}}{\#L_{i}} ;  
    z_{i} = std(t_{i})
    \end{split}
\end{equation}
where $\#L_{i}$ is the size of the list. We reiterate this process on subsequent events until we reach the end of the sample. The above equation transforms strongly active cells into fast cells, and conversely poorly active cells into slow cells. This process is represented in Figure \ref{fig:story_telling}.
A sample $S$ is thus represented by $M$ images $I_1, ..., I_M$ presented successively to the network, which emits a prediction. Each neuron emits at most one spike per image $I_{s}$. The prediction of the network is stored for each image $I_s$, and, at image $I_M$, the pseudo-probability that the sample is of class $c$ is computed given the past information with a temporal softmin:
\begin{equation}\label{eq:softmin}
    P( S = c | I_1, I_2, ..., I_M ) = \pi_{c} = \frac{ \sum_{{s=1}}^{M} e^{ - z_{c}[I_{s}] } }{ \sum_{s=1}^{M} \sum_{j=1}^{K} e^{ - z_{j}[I_{s}] } } 
\end{equation}
where $z_{c}[I_{s}]$ represents the standardized activation time for the $c$-th output neuron at input image $I_s = I_1, ..., I_M$. For audio-based datasets, we first applied the \emph{speech2spike} \citep{speech2spikes} algorithm to transform the raw audio into events, and then applied our \emph{event2time} algorithm.
\subsection{Backward pass}
\label{sec:backward}
In order to train our network, we use the traditional backpropagation algorithm, where gradient descent is performed on the signed-delays $d_{ij}^{s}$. Our aim is to decrease important delays, instead of increasing important weights. The classic learning rule of backpropagation is used:
\begin{equation}\label{eq:backprop}
    d^{s}_{ij} \leftarrow d^{s}_{ij} - \eta \frac{\partial L}{\partial d^{s}_{ij}}
\end{equation}
where $0 < \eta < 1$ is the learning rate. Note that, in practice, we used the Adam optimizer, whose gradient formula is slightly different but can be found in \citep{adam}. The loss function $L$ at the end of the sequence is the traditional cross-entropy loss for classification tasks. And we have:
\begin{equation*}
\label{eq:gradient}
    \frac{\partial L}{ \partial d_{ij}^s } = 
    \frac{\partial L}{\partial \pi_{j}} 
    \frac{\partial \pi_{j}}{\partial z_{j}} 
    \sum_{s=1}^{M} 
    \frac{\partial z_{j}[I_s]}{\partial t_{j}[I_s]} 
    \frac{\partial t_{j}[I_s]}{\partial d_{ij}} 
    \frac{\partial d_{ij}}{\partial d_{ij}^{s}}  
\end{equation*}
where every term is well defined and is derived from differentiable functions, which allows us to directly use generic libraries for automatic differentiation, such as PyTorch \citep{pytorch}. More details on the computation of the gradient can be found in Appendix  \ref{sec:appendix_backward}. 
\subsection{Short and Long Term Memory}
\label{sec:memory}
\noindent Our preprocessing algorithm captures the temporal dynamics of each sample over short windows of time.
Consider that, on average, each input cell presents an activation (or an event) every $\Delta m$ timesteps. If the dataset has $N$ input cells, and we want $2rN$, active cells on each image $I_{s}$, then on average, each image $I_{s}$ represents a short-term window of $\Delta m 2 r N$ timesteps, and the network acquires short term memory. However, for some data, there is a need for longer term memory. Longer term memory is assigned to each neuron of the network as follows: 
\begin{equation}
    \label{eq:delta_eq_signed_longterm}
    \begin{split}
        \delta^{h} &= z_{j}[s] - z_{j}[s-h], \; h = s-\nu, ..., s \\
        z_{j}[s] &\leftarrow z_{j}[s] + \sum_{h=s-\nu}^{s} \alpha_{j}^{h} sign(\delta^{h}) [\exp(-|\delta^{h}|)-1]
    \end{split}
\end{equation}
where $\alpha_{j}^{h} \in [-1, 1]$ is a learnt parameter for neuron $j$. Hence, each cell has a long-term memory equals to $\nu \Delta m 2 r N$, with $\nu$ a constant hyperparameter. More details can be found in Section \ref{sec:long_term_memory}.
\section{Results}
\subsection{Performance on benchmark datasets}
\noindent To show that our network is able to tackle temporal data, we confronted it to the event-based version of MNIST dataset (N-MNIST \citep{nmnist}), to the DVS Gesture dataset \citep{gesture} which represents hand movements, and to the Google Speech Command (GSC) dataset \citep{gsc}, which is a speech recognition dataset. In order to allow for larger comparisons with other models, we also confronted our model to the MNIST \citep{mnist} and CIFAR-10 \citep{cifar10} datasets. Every training details and parameter values can be found in Section \ref{sec:parameter-values}. Results and comparison to the state of the art models are shown in Table \ref{tab:comparison_sota}. For each dataset, we compare to models that either have the best performance or the best accuracy to parameter ratio (see Table \ref{complexities} for more comparisons on performance). While performances are preserved for visual tasks, DeNN improves the performances in audio tasks, where temporality is important. These performances are achieved with architectures much lighter than for other models.  Furthermore, when possible, we computed the average number of computations per sample in different models, by multiplying the reported firing rate to the number of synapses (see Appendix \ref{suppseq_complexities} for more details). We show that it is possible to achieve good performances with fewer computations for MNIST, CIFAR-10, N-MNIST and DVS-Gesture.
\subsection{Energy Consumption}
Based on the theoretical computational complexity of Table \ref{complexities}, energy consumption results have been computed in Table \ref{energy_comp} for the neuromorphic supercomputer of the Human Brain Project, SpiNNaker \citep{painkras2013spinnaker}. We show that DeNN consumes less energy than other best performing models on all datasets, except maybe for DVS gesture, where we could not find the firing rate of the best performing model. Furthermore, as discussed in Section \ref{suppseq_complexities}, these good results will be improved in the near future. Exponential function operations are essential for machine learning. This is leading to increased research in electrical engineering to reduce hardware energy consumption. Impressive power reduction has been achieved recently on the electronic devices used by neuromorphic computers \citep{costa2023exponential}.
\subsection{Choice of kernel $\kappa$}
\label{sec:kernelchoice}
The class of model presented in this work is general enough to work with any kernel, as long as it is a decreasing positive function. Although the kernel that works best is the exponential $\kappa(x) = \exp(-x)$, we found satisfying result with the inverse $\kappa(x)=x^{-1}$. To avoid double negatives (when both $x$ and $d_{ij}^{s}$ are negative), we shifted the Gaussian curve (after standardization) by three units to the right and clipped the (standardized) spike times to a minimum of 0.001. We reported an accuracy of 96.62\% for the MNIST dataset, and 97.76\% for the neuromorphic version. Note that in order to compare only the change in kernel, we used the same architectures and hyperparameters as for $\kappa(x) = \exp(-x)$. It should be possible to obtain even better results by adapting the architectures and parameters of the model. Also, for resources reason, these simulations were performed only on small datasets.
\begin{table}[h!]
\begin{centering}
\caption{Comparison of performance accuracy on benchmark datasets. When possible, we computed the exact number of active synapses in the model. A synapse is considered active if it transmits a value other than 0. See Appendix~\ref{suppseq_complexities} for more details on the average number of computations based on computational complexity. W stands for weight training, and D for delay.}
 \label{tab:comparison_sota}
 \footnotesize
 \centering
  \resizebox{\columnwidth}{!}{\begin{tabular}{ccccc}
    \hline 
    Model & \vtop{ \hbox{\strut \# Parameters} \hbox{\strut / active} } & \vtop{ \hbox{\strut  Avg FLOPS} \hbox{\strut per object} }   &  \vtop{ \hbox{\strut Top-1\%} \hbox{\strut Accuracy} } & \vtop{ \hbox{\strut  Energy cost} \hbox{\strut on SpiNNaker (in $\mu J$)} }   \\
    \hline
    \textbf{MNIST} & & & & \\
    \citep{zhang2020spike}, W & 635,200       &  -       & \textbf{98.40\%} &   $114T$  \\
    \citep{kheradpisheh2020temporal}, W & 317,600 / 35,755 & 35,755 & 97.40\% & 532 \\ 
    DeNN ($q=1$), D            & 79,400 / 14,804         & 14,804          & 97.46\%          &  73  \\
    DeNN ($q=0.5$), D          & 79,400 / \textbf{8,135} & \textbf{8,135}  & 97.43\%          &  $\mathbf{40}$ \\
    \hline
    \textbf{CIFAR-10}  & & & & \\
    \citep{zhou2021temporal}, W & $54.2 \cdot 10^6$                            & $249.4 \cdot 10^{6}$        & \textbf{92.68}\% & 620,190  \\
    \citep{park2021training}, W & $33.6 \cdot 10^6$ & - & 91.90\% & - \\
    DeNN ($q=1$), D             & $5.8 \cdot 10^6$ / $\mathbf{2.3 \cdot 10^6}$ & $61.5 \cdot 10^{6}$     & 90.59\%          & 381,371 \\
    DeNN ($q=0.5$), D           & $5.8 \cdot 10^6$ / $\mathbf{1.4 \cdot 10^6}$ & $\mathbf{43.2 \cdot 10^6}$ & 87.09\%          &  \textbf{232,161}\\
    \hline
    \textbf{N-MNIST} & & & & \\
    \citep{fang2021incorporating}, W& -           & -                   & \textbf{99.61\%} & $13,687\tau + 375$  \\ 
    \citep{zhu2022eventbackprop}, W & 35,800      & -                   & 99.39\% & $4301(0.9\tau +1)$ \\
    DeNN ($q=0.5$), D             & 15,696 / \textbf{11,788} & \textbf{665,262} & 98.06\% & $\mathbf{11,616}$    \\
    \hline
    \textbf{DVS-Gesture} & & & & \\
    \citep{cordone2021learning}, W  & 13,992           & $< 10 \cdot 10^6$ & 92.01\%   &  9.6383        \\
    \citep{yao2023attention}, W& -                & -                 & \textbf{98.23\%}  & -    \\  
    DeNN ($q=0.5$), D             & 19,216 / \textbf{7,895} & $\mathbf{5.8 \cdot 10^6}$ & 97.57\%   & 312,476    \\
    \hline
    \textbf{GSC} & & & & \\  
    \citep{bittar2024oscillations}, W+D & $1.5 \cdot 10^{6}$ & - &  97.05\% & $39,481(\tau + 0.21)$ \\
    \citep{deckers2024delays}, W+D  & 610,000 & $\boldsymbol{\sim 3.45 \cdot 10^{6}}$ & 95.69\% & 25,005  \\
    DeNN ($q=1$), D                 & \textbf{175,467} & $\sim 3.6 \cdot 10^{6}$ & \textbf{97.73\%} & $\mathbf{20,715}$    \\
    \hline
 \end{tabular} }
 \end{centering}
\end{table}
\subsection{Time for event-based models and event-based datasets}
\begin{figure}[h!]
    \centering 
    \includegraphics[width=0.8\linewidth]{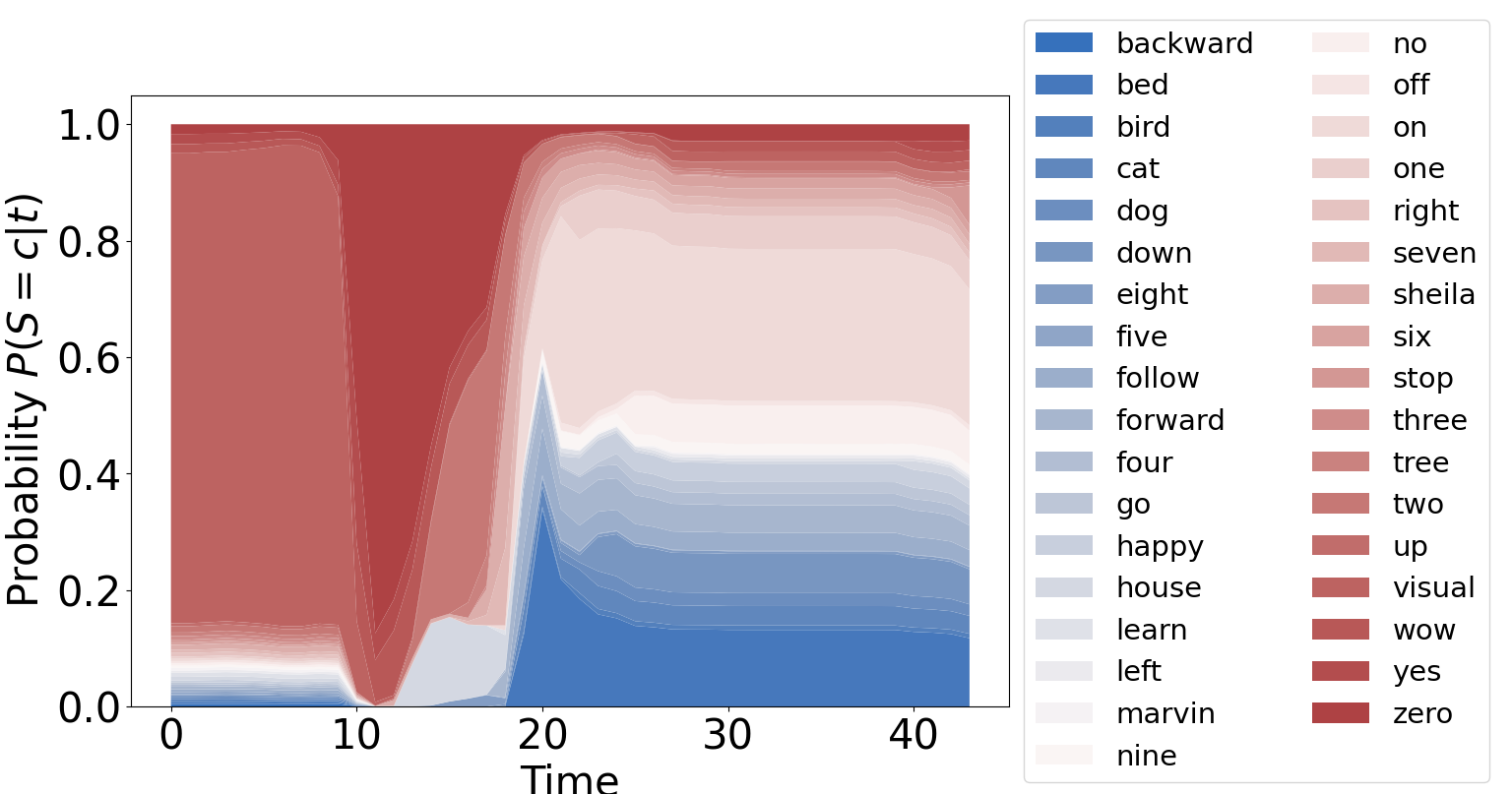}
    \caption{Graph of probabilities $p$ for each class of the dataset given the past at each timestep $t$, for a sound of the GSC dataset ("Off").}
    \label{fig:probs}
\end{figure}

\begin{figure}[h!]
    \begin{subfigure}[t]{0.49\textwidth}
        \centering
        \includegraphics[width=\linewidth]{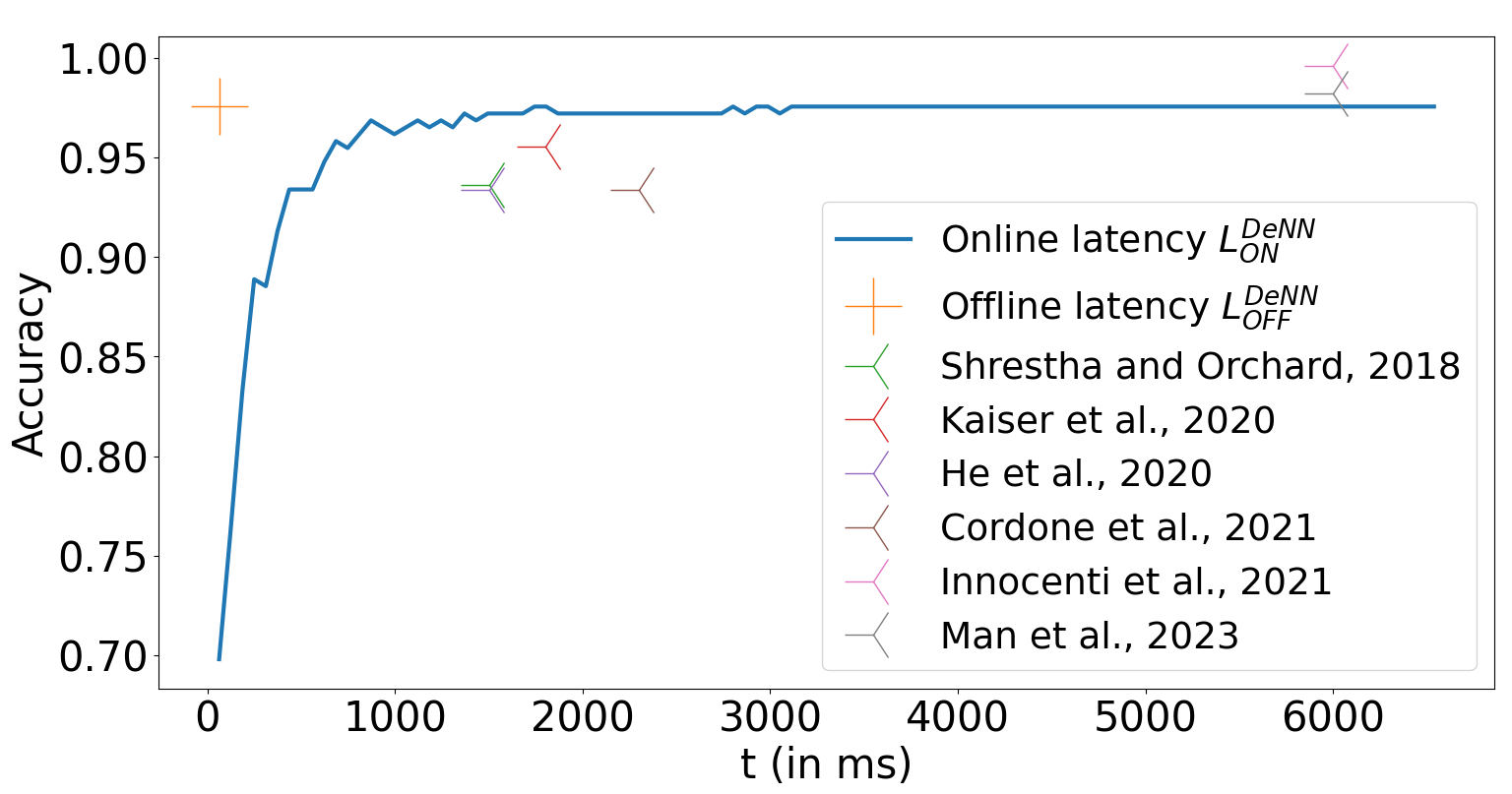}
        \caption{Accuracy obtained with different maximum timesteps for a model trained on full samples, for the DVS-Gesture dataset.}
        \label{fig:truncated_t}
    \end{subfigure} 
    \hfill
    \begin{subfigure}[t]{0.49\textwidth}
    \centering
        \includegraphics[width=\linewidth]{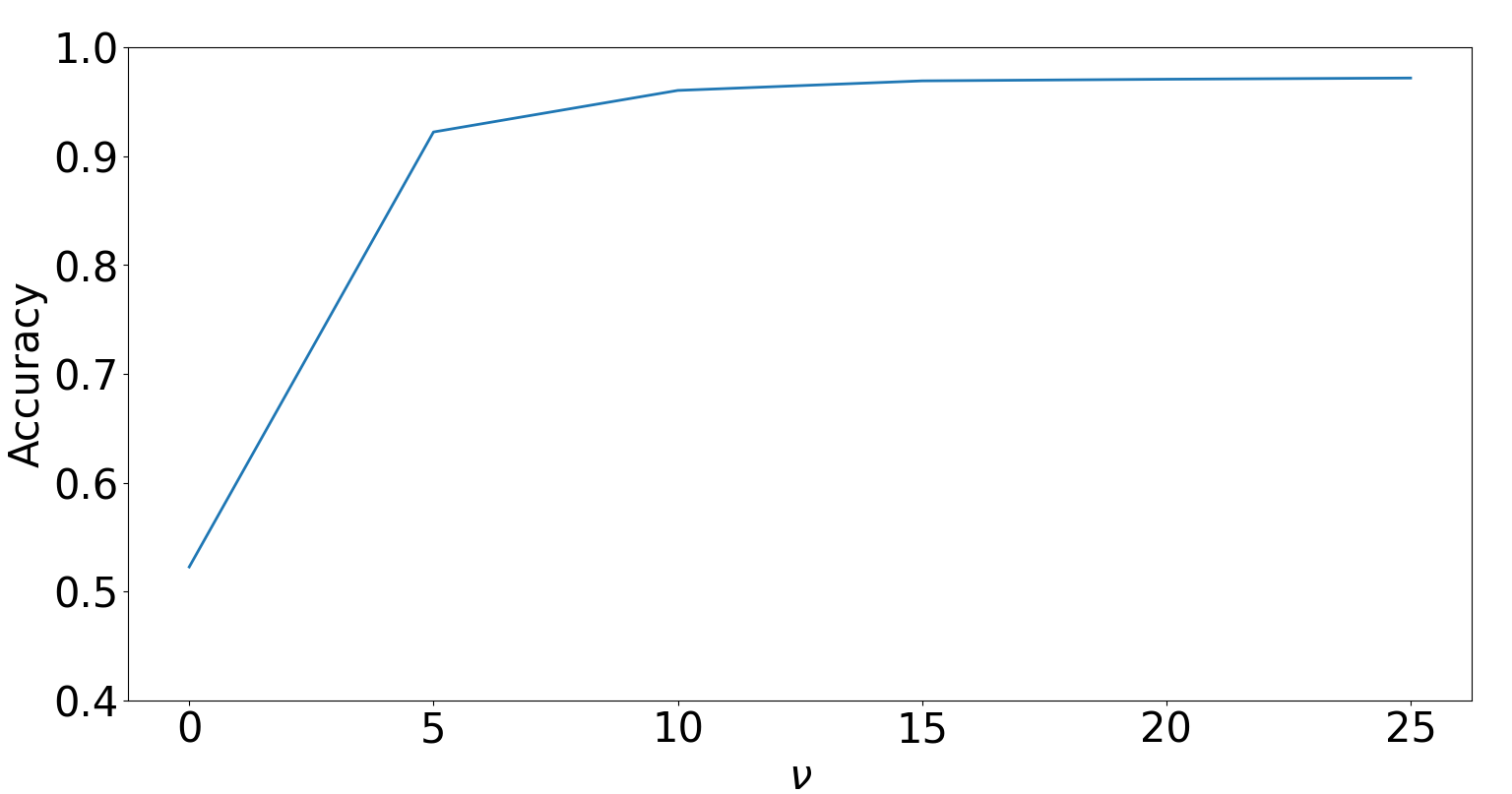}
        \caption{Accuracy obtained on the GSC dataset with models with short ($\nu=0$) to long-term memory.}
        \label{fig:nu_accuracy}
    \end{subfigure}
    \caption{Ablation studies for DVS-Gesture and GSC}\vspace{-0.3cm}
    \label{fig:temporal_use}
\end{figure}

\begin{figure}[h!]
    \centering 
    \includegraphics[width=0.8\linewidth]{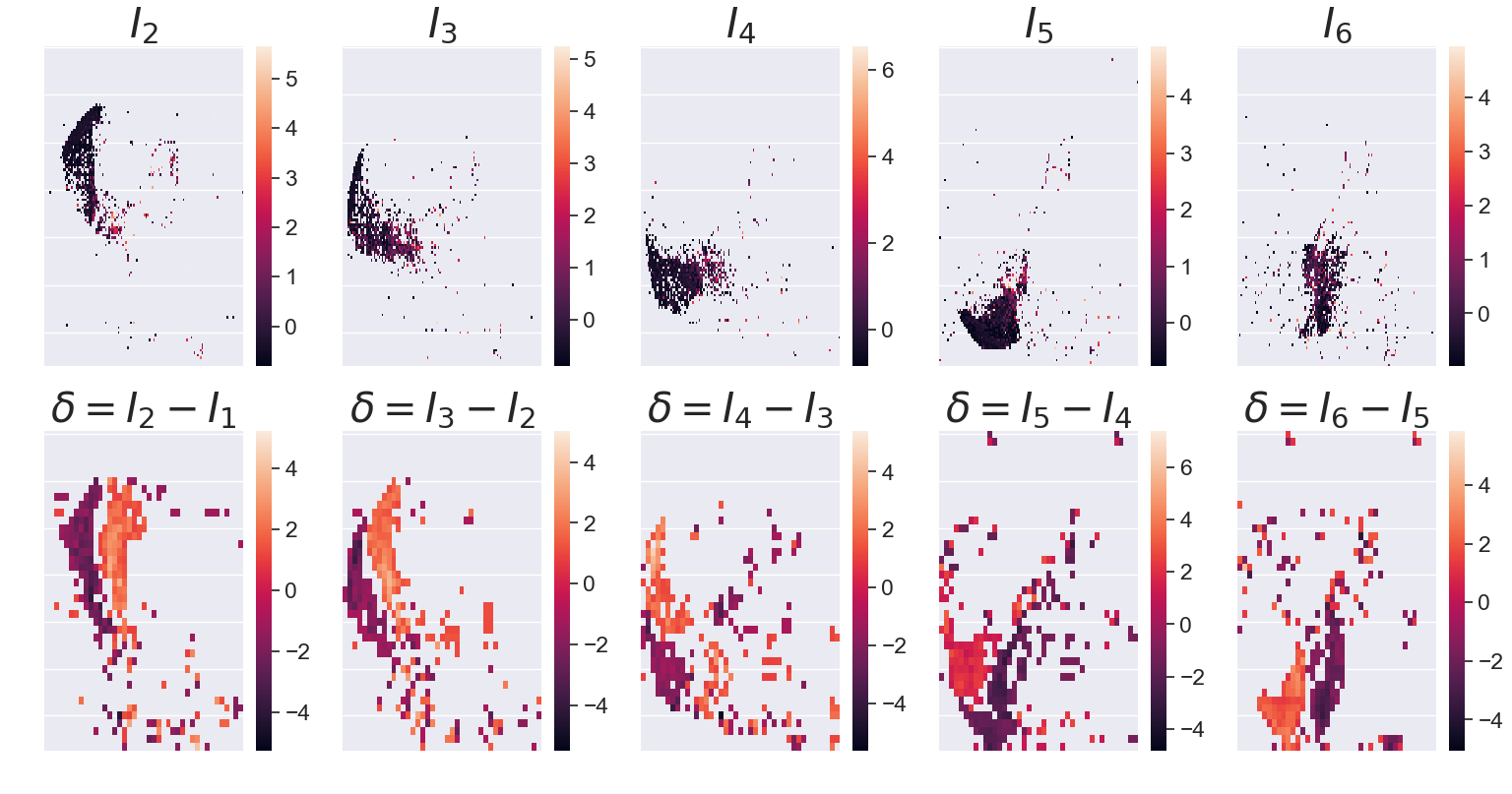}
    \caption{Top line: Input images $I_{2}$ to $I_{6}$ from a right-hand counter clockwise movement of the DVS-Gesture dataset, obtained after application of our preprocessing algorithm. Bottom line: Differences $\delta$ between neurons of the first convolutional layer's feature maps at images $I_{s}$ and $I_{s-1}$. Darker pixels indicate faster neurons between two timesteps. Note that neurons in the range $[-1, 1]$ are canceled for clarity of image, leaving us with neurons where the difference is significant enough.}
    \label{fig:movement}
\end{figure}

To show how a DeNN uses the temporal information in event-based datasets, Figure \ref{fig:probs} depicts the probabilities that a sound $S$ is of class $c$ given the past, for a sample of sound drawn from the GSC dataset. The probabilities clearly evolve with time and inputs, as the networks get more information about the stimulus. Figure \ref{fig:nu_accuracy} shows the evolution of accuracy on GSC dataset for different values of long-term memory. With hyperparameter memory length value $\nu=5$ (Equation \ref{eq:delta_eq_signed_longterm}), the model already classifies correctly about 93\% of inputs, and it crosses 96\% at $\nu \geq 10$.
We also note that the activity of neurons accelerates in the direction of the movement for the DVS-Gesture dataset, as illustrated in Figure \ref{fig:movement}. We computed the difference $\delta$ (see Equation \ref{eq:delta_eq_signed_longterm}) between neurons of the first layer's feature maps created after the input of a sample of the DVS-Gesture dataset. We found that, in the direction of a movement (here a right-hand counter clockwise), neurons tend to decrease their relative timing spike to others (i.e. $z_{j}[s]<z_{j}[s-1]$), while they increase afterwards. This is consistent with the hypothesis that time in visual system encodes speed and direction of stimuli \citep{vahdatpour2024latency}. To further explore the time dimension in the DVS-Gesture dataset, an experiment was run where the total number of timesteps during inference was truncated. The results are presented in Figure \ref{fig:truncated_t}. It is shown that a DeNN can achieve good performance in a few timesteps, which hints that the time dimension in the DVS-Gesture dataset might not be the most important dimension. Indeed, samples in this dataset are composed of periodic movements. Hence, only a few periods are required to really discriminate between movements.
\subsection{Latency}
Figure \ref{fig:story_telling} shows that while SNN requires a discrete time simulation in real time, with a $ \Delta t $ step (see step (1)), DeNN can use a discrete event simulation without any $ \Delta t $ step (see step (3)). As shown in step (2), DeNN on-line latency is theoretically lower than SNN on-line latency, because a DeNN does not have to wait for the last $ \Delta t $ step to be consumed after last image at $ T \Delta t $ step. For the online latency, a timestep on the Gesture dataset represents, on average, 0.6227 seconds of the sample, and 0.6099 seconds for the N-MNIST dataset. On-line property relies on the fact that input events are received in real time (e.g., by an event-based camera of a robot/car). However, as shown in step (3), if on-line learning in real time is not required, DeNN allows as fast as possible discrete event simulation. All the images are fed to the network in a sequence leading to smaller off-line latency $L^{DeNN}_{OFF}= 0.066$ seconds on a NVIDIA GeForce RTX 3080 Laptop GPU; for the Gesture dataset. This is particularly interesting for example to train cars for automated driving by off line simulation. This allows greatly reducing training time and energy consumption.
We show on Figure \ref{fig:truncated_t} the comparison with other SNN models for the Gesture dataset. We note that models reporting better accuracy are also models with higher latencies.
Note that this as-fast-as-possible discrete event simulations do not depend on slow or fast regime. Although the slow regime requires receiving all the spikes from the previous layer to compute the output spike times of the layer, it does not require waiting for any real time discrete time step $\Delta t$. Even in a slow regime, layers can be simulated as fast as possible. The only difference of the fast regime is that it reduces the number of computations and execution time, while slightly decreasing performances.
\subsection{Functioning of the network}
\begin{figure}[h!]
    \centering
    \includegraphics[width=0.80\linewidth]{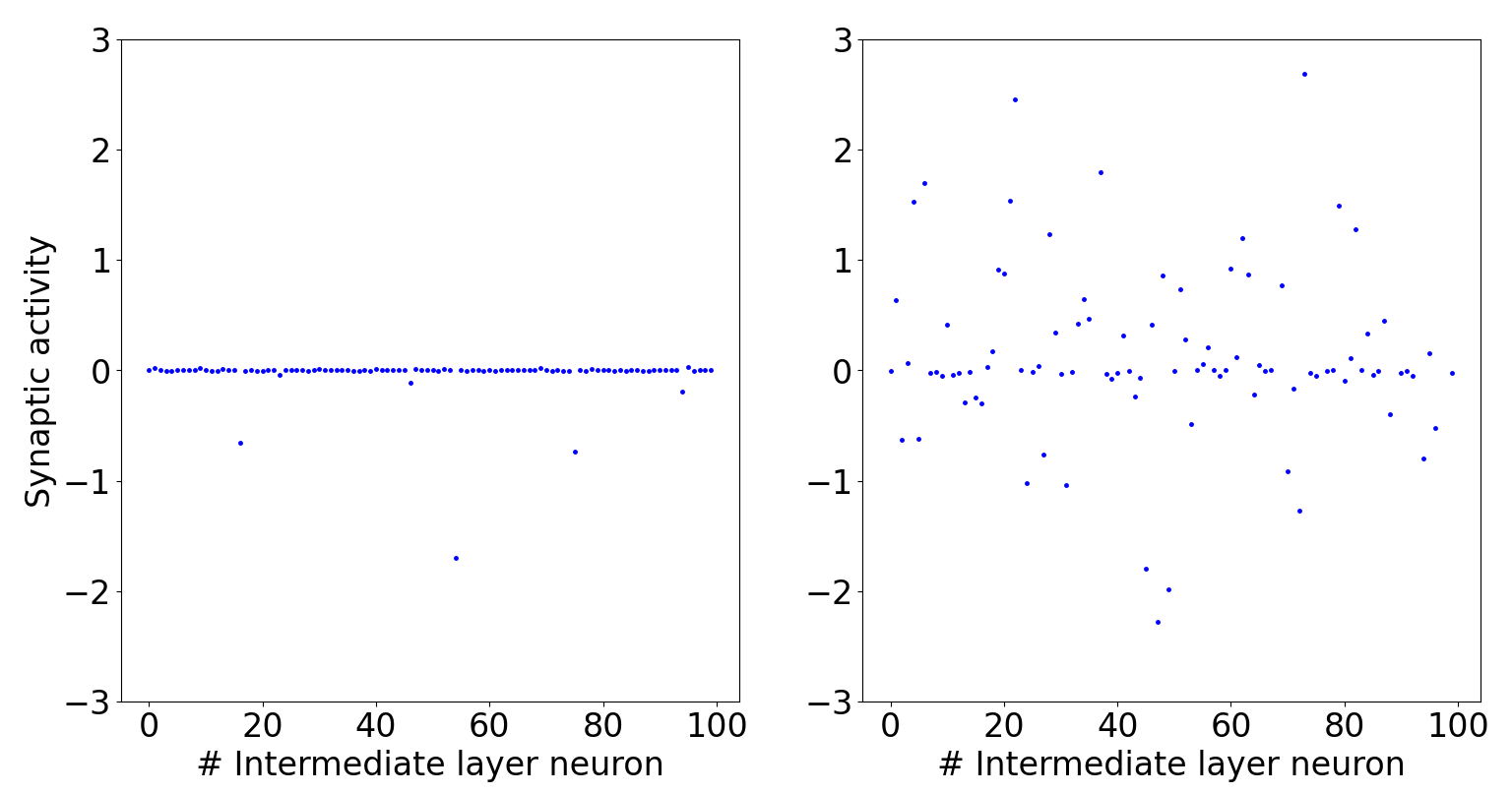}
    \vspace{-0.3cm}\caption{Synaptic activity is defined as $sign(d_{ij}^{s})[\kappa(z_{i}+d_{ij})-\kappa(z_{i}+1)]$  for a DeNN(on the left) and $w_{ij}x_{i}$ for an ANN (on the right). Each dot is a synapse.}
    \label{fig:synaptic_impact}
\end{figure}

Figure \ref{fig:synaptic_impact} shows the activity of synapses connecting the neurons in the intermediate layer of the classifier to the eighth output neuron, for the MNIST dataset, for images of the digit 8. In a DeNN (left), very few synapses are of extreme importance to make the output neuron spike much before the others, while others synapses are silent. In an ANN (right), there must be a balance between inhibition and excitation. The main difference between the ANN and the DeNN stems in the fact that, in our network, paths that are irrelevant do not show any activity, while in an ANN they are inhibited. This shows how DeNN is able to drastically decrease the number of computations with respect to other models. All couples (neuron, digit) have been computed and are available on figures \ref{matrix_mnist} (for MNIST) and \ref{matrix_cifar} (for CIFAR-10).

This behaviour might be due to the peculiar derivative of $t_{j}$ with respect to the parameter $d_{ij}^{s}$. As shown in Figure \ref{fig:theoretical_derivative}, the derivative is close or equal to zero for almost every input $z_{i}$. Even when the input is sufficiently strong (reminding that negative $z$ is quicker input than others), the derivative plunges toward zero at $d_{ij}^{s} = 0$, which means that synapses will have trouble changing signs. Some will get asymptotically close to zero (negative or positive), while never being able to change sign. 
 \vspace{-0.3cm}
\section{Conclusion}
A new class of neural networks was presented. These networks are able to treat temporal information, arguing they are, alongside \citep{hazan2022memory}, the first network to be fully temporally coded, by taking into account the spike order's importance, as presented in \citep{thorpe2001spike, liu2024firstspike}. Indeed, the model in \citep{hazan2022memory} is temporally coded, but lacks performances. Finally, DeNN are able to achieve satisfying results on tasks with much less parameters than other models, and with less computations. On audio task, where temporality is important, DeNN demonstrates better results than state-of-the-art models.  We also show that DeNN use less energy than other models. On datasets where temporal dimension is crucial, DeNN use less energy with better performance results than the adaptive LIF model used in \citep{bittar2022gradient, bittar2024oscillations, deckers2024delays}.

Since the equations for the DeNN are general, it is possible to adapt any (continuous) architecture to the DeNN with few efforts. We were indeed able to very easily apply our model to convolutional architectures, and try out different kernels $\kappa$ fairly easily. Conversion of existing architectures to spiking one would be an interesting development for DeNN, as in \citep{yao2023attention, yao2024transformer}. A limitation can be discussed for long-term memory hard window definition. The memory cost increases with $\nu$ for sequences longer than a few seconds. However we have to note that thanks to our preprocessing algorithm, one timestep in our model is equal to $\Delta m2rN$ timesteps in real time. With $\nu$, the memory goes to  $\nu \Delta m2rN$, as explained in Section \ref{sec:memory}. It should be possible to encode sequences longer than a few seconds with few adaptations to the preprocessing algorithm (for example, with a bigger $r$). Also, we show in Table \ref{energy_comp} that even with $\nu=25$, the DeNN uses less energy than the adLIF used in \citep{deckers2024delays}, which uses coupled dynamic equations, one for the short memory (membrane potential), and one for the long-term memory (recovery variable).

This work should open new perspectives to the field of temporal coded networks, as the need to better take into account the temporality of information, especially with the rising use of event-based cameras and brain-inspired learning.
\section*{Data and code availability}
\noindent The datasets used in this study are publicly available on the Internet. Code is available from the corresponding author on request.
\section*{Acknowledgements}
\noindent This work was supported by the French government, by the National Research Agency with the DeepSee project ANR-20-CE23-0004 and  by the interdisciplinary Institute for Modeling in Neuroscience and Cognition (NeuroMod) of the Universit\'e C\^ote d'Azur. It is part of the \href{https://explain.i3s.univ-cotedazur.fr}{eXplAIn} CNRS research team. 
\href{https://wiki.inria.fr/ClustersSophia/Usage_policy}{NEF} computing platform from Inria Côte d'Azur has been used for running or parallel simulations. NEF is part of the \href{https://univ-cotedazur.eu/services-for-research-scientists/opal-computing-center}{OPAL} distributed computation mesocentre. The authors are grateful to the OPAL infrastructure from Université Côte d'Azur for providing resources and support.
The authors would like to thank Andrew Rowley, researcher from the Human Brain Project in the School of Computer Science at the University of Manchester, and in charge of SpiNNaker developments, for his profound insights of the energy consumption of the SpiNNaker chip.

\bibliographystyle{chicago} 
\bibliography{refs.bib}

\section*{Supplementary}
\appendix
\counterwithin{figure}{section}
\counterwithin{table}{section}
\renewcommand*\thetable{\Alph{section}.\arabic{table}}
\renewcommand*\thefigure{\Alph{section}.\arabic{figure}}
\section{Spiking time continuity}
\label{sec:continuity}
\begin{figure}[h]
    \centering
    \includegraphics[width=0.7\linewidth]{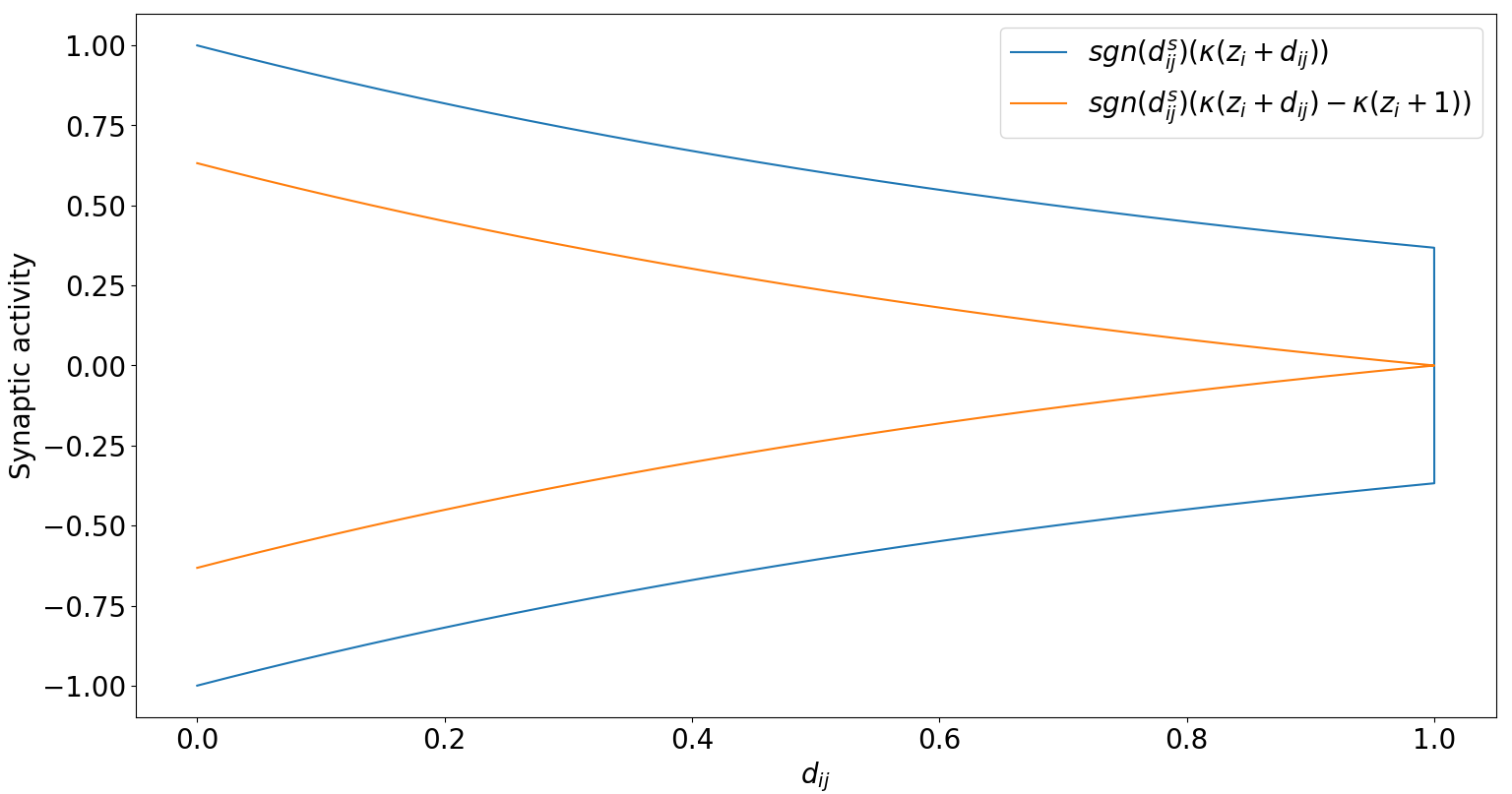}
    \caption{Synaptic activity with and without the correction term, as a function of delays $d_{ij}$.}
    \label{fig:continuity}
\end{figure}

We show on Figure \ref{fig:continuity} how the term $-\kappa(z_{i}+1)$ corrects a discontinuity at $d_{ij}=1$. If $d_{ij} = 1$, $d_{ij}^{s} = 0$.
\begin{equation*}
    \label{eq:continuity}
    \begin{split}
        sgn(d_{ij}^{s}) kappa( t + d_{ij} ) &\xrightarrow[d_{ij}^{s} \to 0^{-}]{} - kappa( t + 1 )\\
        sgn(d_{ij}^{s}) kappa( t + d_{ij} ) &\xrightarrow[d_{ij}^{s} \to 0^{+}]{}  kappa( t + 1 )
    \end{split}
\end{equation*}
Since $\kappa$ is a strictly decreasing positive function, $-kappa( t + 1 ) < kappa( t + 1)$, hence there is a discontinuity at $d_{ij} = 1$.
\section{Spiking time standardization}
\label{sec:standardization}
Standardization process represents a shift in time of the $\kappa$ kernel. Indeed, set the quantile value $t_q$ after which incoming spikes are ignored. Shifting the $\kappa$ kernel, we get:
\begin{equation*}
    t_j = \sum_i sign(d_{ij}^{s}) \Big[ \kappa(t_{i} + d_{ij} - t_q) - \kappa(t_{i} + 1 - t_q) \Big]
\end{equation*}
which is almost equivalent (up to a division by standard deviation) to the standardization process. Taking the median quantile $t_{q} = t_{0.5}$, if $t_{i}$ are normally distributed (see Figure \ref{fig:neurons_before_std}), we have:
\begin{equation*}
\begin{split}
        z_{i} + d_{ij} = \frac{t_{i}-t_{0.5}}{\sigma} + d_{ij} &\sim N(d_{ij},1)\\
        t_{i} - t_{0.5} + d_{ij} &\sim N(d_{ij}, \sigma).
\end{split}
\end{equation*}
The difference we get in evaluating $\kappa( t_{i} + d{ij} - t_{0.5} )$ and $\kappa( \frac{t_{i}-t_{0.5}}{\sigma} + d_{ij} )$ depends on $\sigma$. It can get high the further $\sigma$ gets from 1. However, dividing by $\sigma$ allows for better numerical stability in the evaluation of $\kappa$ kernel, because it allows every layer to operate in the same range of values. Moreover, the division will reflect during backpropagation algorithm, because it will appear in the derivatives. Hence, the two mechanisms are ``almost'' equivalent.
\section{Backward computation}
\label{sec:appendix_backward}
\begin{figure}[h]
    \centering
    \includegraphics[width=\linewidth]{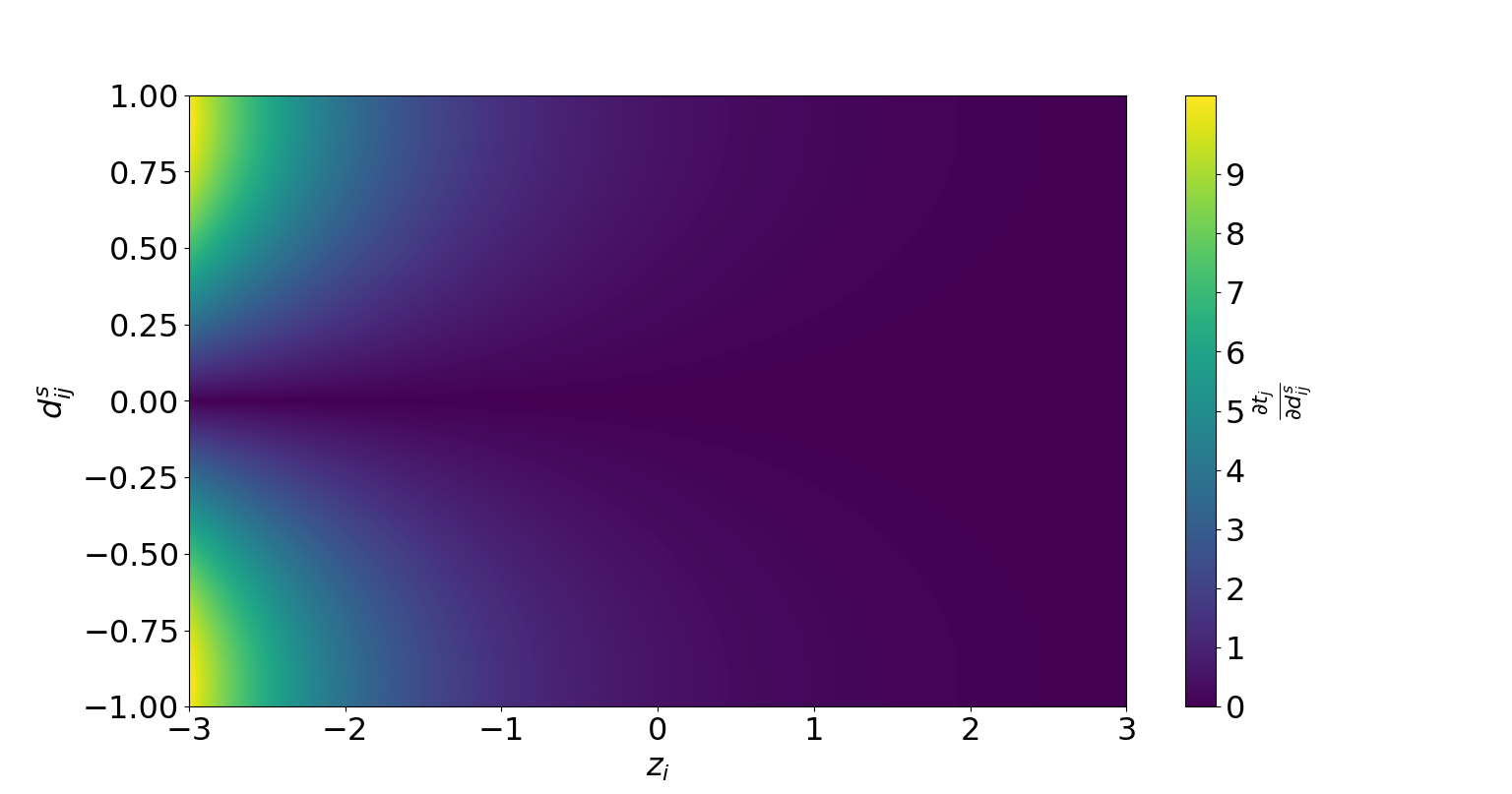}
    \caption{Derivative of $t_{j}$ with respect to parameter $d_{ij}^{s}$.}
    \label{fig:theoretical_derivative}
\end{figure}

The loss function at the end of the sequence is the traditional cross-entropy loss for classification tasks:
\begin{equation*}
    L = - \sum_c^K target_c \log(\pi_{c})
\end{equation*}
with
\begin{equation*}
    P( S = c | I_1, I_2, ..., I_T ) = \pi_{c} = \frac{ \sum_{{s}}^{T} e^{ - z_{c}[I_{s}] } }{ \sum_{s}^{T} \sum_{j}^{K} e^{ - z_{j}[I_{s}] } } 
\end{equation*}
and
\begin{equation*}
    \frac{\partial L}{\partial \pi_{c}} = - \frac{1}{\pi_{c}}
\end{equation*}
\begin{equation*}
  \frac{\partial \pi_{c}}{z_{l}} =
    \begin{cases}
      \pi_{c} ( \pi_{c} - 1 )    & \text{if $l=c$}\\
      \pi_{c}\pi_{l} & \text{if $l\neq c$}
    \end{cases}       
\end{equation*}
For differentiating the $z_{j}$ variable, we need to take into account the fact that $z_{j}$ depends on the input $I_{s}$. Thus:
\begin{equation*}
    \frac{\partial z_{j}}{\partial d_{ij}^{s}} = 
    \sum_{s}^{T} 
    \frac{\partial z_{j}[I_s]}{\partial t_{j}[I_s]} 
    \frac{\partial t_{j}[I_s]}{\partial d_{ij}} 
    \frac{\partial d_{ij}}{\partial d_{ij}^{s}} 
\end{equation*}
with $\kappa(x) = e^{-x}$ as kernel, we have:
\begin{equation*}
    \frac{\partial t_{j}}{\partial d_{ij}^{s}} [I_{s}] =  \frac{2|d_{ij}^{s}|}{\sigma^2} d_{ij} e^{- ( z_{i}[I_{s}] + d_{ij} )  }
\end{equation*}
represented on Figure \ref{fig:theoretical_derivative}. In order to tackle vanishing and exploding gradients which could arise, a gradient normalization is implemented at each layer, using the Frobenius norm of the gradient matrix.
\section{Training details}
\label{sec:parameter-values}
Widespread PyTorch library (Python version 3.8.6, PyTorch 1.10.0, \citep{pytorch}) has been used for achieving fair performance comparisons and also for its high capabilities. Experiments were conducted on a RTX 2080 Ti GPU. 

Models' architectures are described in Table \ref{archi} and parameters used for learning on each dataset are detailed in Table \ref{tab:parameters} in Appendix. The Adam algorithm \citep{adam} with default parameters is used to perform the backpropagation.
\begin{table}[h!]
\centering
\caption{Architectures used for the different datasets. A convolutional layer described as 8c5s2 means 8 filters of size 5x5 with a stride of 2. A minpool layer with 2x2 kernels and a stride of 2 is described as p2s2.}
\label{archi}
\begin{tabular}{cc}
Dataset     & Model Architecture                                                                                 \\ \hline
MNIST       & 784-100-10                                                                                         \\
CIFAR-10    & VGG-9 \\
N-MNIST     & 8c5s2 - 16c3s1 - p2s2 - 32c3s1 - 32c3s1 - p2s2 - 10                                                \\
DVS-Gesture & 8c7s3 - 16c5s2 - p2s2 - 32c3s1 - 32c3s1 - p2s2 - 11   \\
GSC & 60-256-256-256-35
\end{tabular}
\end{table}
\begin{table}[h!]
\begin{center}
\caption{Parameter values for each dataset.}
\label{tab:parameters}
\begin{tabular}{c|ccccc}
Parameters           & MNIST & CIFAR-10 & N-MNIST & DVS-Gesture & GSC              \\ 
\hline
Batch size           & 4096  & 512 & 16      & 16          & 700              \\
Learning Rate        & 0.001 & 0.001 & 0.001   & 0.001       & 0.001            \\
LR Scheduler         & - & - & -       & -           & CosineAnnealing  \\
$\Delta$             & - & - & 4       & 4           & 1                \\
$r$                  & - & - & 0.05    & 0.05        & 0.1              \\            
Seed                 & 22756400 & 76446569 & 94240977& 98074194    & 36887311         \\
s2s channels         & - & - & -       & -           & 30               \\
s2s threshold        & - & - & -       & -           & 0.75  \\
$\nu$                & - & - & 0       & 0           & 25               \\      
\end{tabular}
\end{center}
\end{table}
\section{Long-term memory}
\label{sec:long_term_memory}

\begin{figure}[h]
    \centering
    \includegraphics[width=0.8\linewidth]{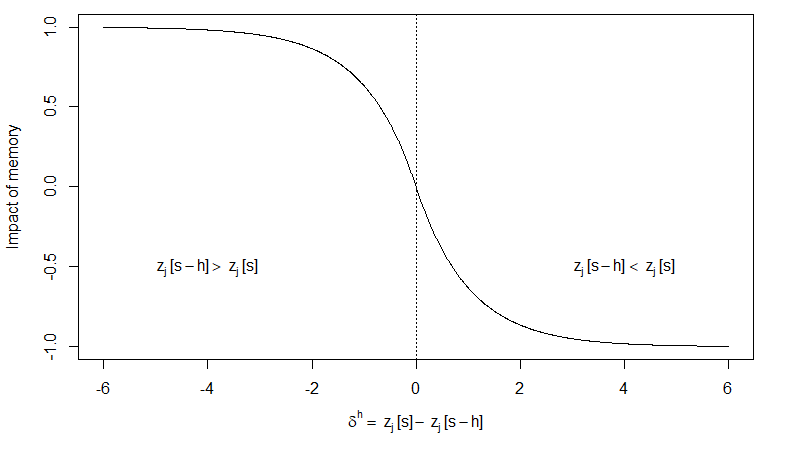}
    \caption{Representation of the term $sign(\delta^{h}) [\exp(-|\delta^{h}|)-1]$ in Equation \ref{eq_supp:delta_eq_signed_longterm}}
    \label{fig_supp:deltah}
\end{figure}

Long-term memory is added to the network thanks to this equation:
\begin{equation*}
    \label{eq_supp:delta_eq_signed_longterm}
    \begin{split}
        \delta^{h} &= z_{j}[s] - z_{j}[s-h], \; h = s-\nu, ..., s \\
        z_{j}[s] &\leftarrow z_{j}[s] + \sum_{h=s-\nu}^{s} \alpha_{j}^{h} sign(\delta^{h}) [\exp(-|\delta^{h}|)-1]
    \end{split}
\end{equation*}
The mechanism is represented on Figure \ref{fig_supp:deltah}. The main mechanism is that, if neuron $j$ is faster than its neighbors ($\delta^{h}>0$ or $z_{j}[s-h]<z_{j}[s]$), then it gets a small boost ($z_{j}[s]$ is decreased). Conversely, if it was slower than its neighbors, then it gets a small punishment ($z_{j}[s]$ is increased). Note that $\alpha_{j}^{h}$ is a learnable parameter in the range $[-1,1]$, so the network can decide to reverse boost and punishment.
\section{Computational complexities \& energy cost}
\label{suppseq_complexities}
It is described here how to obtain the average number of computations per object derived for Table \ref{tab:comparison_sota}.

For sparse convolutions used in \citep{cordone2021learning}, the average number of active sites $n_a$ \citep{graham2018sc} per layer per timestep can be computed with the ratio of the number of spikes in a layer to the number of timesteps. However, we are unable to know the spatial distribution of activated sites. For a lower bound estimation, it can be assumed that each of them was visited exactly once, thus underestimating the number of computations. The lower bound for average number of computations per layer and timestep is then $n_a C^{\ell}_{o} C^{\ell-1}_{o}$, with $C^{\ell}$ the number of channels of the convolutional layer, and the total number of computations per one sample of the dataset is obtained by summing over all layers and all timesteps, and is approximately of $1.2 \cdot 10^6$ computations. An upper bound can be obtained by assuming that active sites are densely grouped on a square of side $\sqrt{n_a}$, which gives, when summing over all layers and timesteps, approximately $10 \cdot 10^6$ computations. 


\begin{table}[h!]
  \caption{We set $n^{\ell}$ the number of neurons in layer $\ell$,  $n^{\ell}_{s}$ the number of spiking neurons in layer $\ell$, while $T$ is the total number of timesteps. $I$ and $C$ represent sets of causal spikes \citep{mostafa2017supervised, goltz2021fast, comsa2020temporal}. For rate coding networks, $\tau$ represents the ratio of the total number of spikes observed in $T$ timesteps to the total number of neurons \citep{datta2021training}. For convolutional layers, $H^{\ell}_{o}, W^{\ell}_{o}, C^{\ell}_{o}$ and $k^{\ell}$ are the height and width of the feature maps, the number of channels and the size of the kernel.}
  \label{complexities}
  \centering
  \footnotesize
  \resizebox{\columnwidth}{!}{\begin{tabular}{cccc}
    \hline
    \hbox{\strut Dataset} \\
    \hbox{\strut
    Neural coding
    }& Model   & \vtop{ \hbox{\strut  Computational} \hbox{\strut complexity} } & Accuracy   \\
    \hline
    \textbf{MNIST} & &   \\
    TTFS &\citep{zhang2020spike}    & $\mathcal{O}\big(n^{l}[T-1+T(n_{s}^{\ell-1}+1)] \big)$ & 98.40\%   \\
    TTFS &\citep{kheradpisheh2022bs4nn}  & $\mathcal{O}\big( n^{\ell} [ T-1 + T(n^{\ell-1}_{s}+1) + n^{\ell-1}_{\bar{s}} ] \big)  $ & 97.00\% \\
    TTFS&\citep{comsa2020temporal}       & $ \mathcal{O}\big( n^{\ell-1} + |I|n^{\ell} \big) $    & 97.96\%   \\
    TTFS&\citep{im2022chip} & $\mathcal{O}\big( n^{\ell} [T-1 + T(n^{\ell-1}_{s}+1)] \big)  $    & 96.00\%  \\
    TTFS&\citep{oh2022circuits}            & $ \mathcal{O}\big( T[ n^{\ell} ( n^{\ell-1}_{s} + 1 ) ] \big) $  & 96.90\%   \\
    TTFS&\citep{mostafa2017supervised}  & $ \mathcal{O}\big( n^{\ell-1} + 2|C|n^{\ell} \big) $   & 97.55\%    \\
    TTFS&\citep{goltz2021fast}     & $ \mathcal{O}\big( n^{\ell-1} + |C|n^{\ell} \big) $  & 97.10\%        \\
    TTFS&\citep{kheradpisheh2020temporal}   & $\mathcal{O}\big( n^{\ell} [T(n^{\ell-1}_{s}+1) + T-1 + n^{\ell-1}_{\bar{s}}] \big)  $   & 97.40\% \\
    Temporal&DeNN ($q=1$)  &  $ \mathcal{O}\big( n^{\ell}[ n^{\ell-1}+2 ] \big) $ & 97.46\%\\
    Temporal&DeNN ($q=0.5$)  & $ \mathcal{O}\big( n^{\ell}[ n_{s}^{\ell-1}+2 ] \big) $ & 97.43\%    \\
    \hline
    \textbf{CIFAR-10} & &  \\
    TTFS &\citep{datta2021training}       & $\mathcal{O}\big(C^{\ell}_{o} H^{\ell}_{o} W^{\ell}_{o} [ \tau C^{\ell-1}_{o} C^{\ell}_{o} k^{\ell^2} + 2T -1 ]\big)$ & 91.41\% \\
    TTFS &\citep{zhou2021temporal}    & $\mathcal{O}\big( C^{\ell-1}_{o} k^{\ell^2} [ H^{\ell}_{o} W^{\ell}_{o} k^{\ell^2} (|C|+1) + C^{\ell}_{o} H^{\ell}_{o} W^{\ell}_{o}]  \big)$  & 92.68\%  \\
    TTFS&\citep{park2020t2fsnn}           & $\mathcal{O}\big(  H^{\ell}_{o} W^{\ell}_{o} C^{\ell}_{o} [ \tau C^{\ell-1}_{o} k^{\ell^2} + 1 ]  \big)$  & 91.43\%   \\
    Temporal&DeNN ($q=1$)      & $\mathcal{O}\big( C^{\ell}_{o} H^{\ell}_{o} W^{\ell}_{o} [C^{\ell-1}_{o} k^{\ell^2} +2]  \big)$    & 90.59\%   \\
    Temporal&DeNN ($q=0.5$)      & $\mathcal{O}\big( C^{\ell}_{o} H^{\ell}_{o} W^{\ell}_{o} [\tau C^{\ell-1}_{o} k^{\ell^2} +2]  \big)$ & 87.09\%      \\
    \hline
    \textbf{N-MNIST} & &  \\
    Rate &\citep{lee2016training}     & $\mathcal{O}\big( n^{\ell} [ \tau n^{\ell-1} + T ( 2  + n^{\ell} )] \big)$  & 98.74\%    \\
    Rate&\citep{wu2018spatio}         & $\mathcal{O}\big( n^{\ell} [ \tau n^{\ell-1} + T ] \big)$      & 98.78\%       \\
    Rate&\citep{shrestha2018slayer}   & $\mathcal{O}\big( H^{\ell}_{o} W^{\ell}_{o} C^{\ell}_{o} [ T + \tau C^{\ell-1}_{o}k^{\ell^2} ] \big)$   & 99.20\%           \\
    Rate&\citep{kaiser2020synaptic}   & $\mathcal{O}\big( H^{\ell}_{o} W^{\ell}_{o} C^{\ell}_{o} [ T + \tau C^{\ell-1}_{o}k^{\ell^2} ] \big)$    & 96\%           \\
    Rate&\citep{jin2018hybrid}        & $\mathcal{O}\big( n^{\ell} [ \tau n^{\ell-1} + T ] \big)$   & 98.84\%           \\
    Rate&\citep{lee2020enabling}      & $\mathcal{O}\big( H^{\ell}_{o} W^{\ell}_{o} C^{\ell}_{o} [ T + \tau C^{\ell-1}_{o}k^{\ell^2} ] \big)$   & 99.09\%                    \\
    Rate&\citep{cheng2020lisnn}       & $\mathcal{O}\big( H^{\ell}_{o} W^{\ell}_{o} C^{\ell}_{o} [ \tau ( C^{\ell-1}_{o}k^{\ell^2} + k_{\omega}^{2} ) + T)  ] \big)$              & 99.45\%        \\
    Rate&\citep{fang2020exploiting}   & $\mathcal{O}\big( H^{\ell}_{o} W^{\ell}_{o} C^{\ell}_{o} [ T + \tau C^{\ell-1}_{o}k^{\ell^2} ] \big)$  & 99.39\%                  \\
    Rate&\citep{he2020comparing}      & $\mathcal{O}\big( n^{\ell} [ \tau n^{\ell-1} + T ] \big)$   & 98.28\%                \\
    Rate&\citep{fang2021incorporating}& $\mathcal{O}\big( H^{\ell}_{o} W^{\ell}_{o} C^{\ell}_{o} [ T + \tau C^{\ell-1}_{o}k^{\ell^2} ] \big)$   & 99.61\%            \ \\ 
    Temporal&DeNN ($q=0.5$)                   & $\mathcal{O}\big( C^{\ell}_{o} H^{\ell}_{o} W^{\ell}_{o} [\tau C^{\ell-1}_{o} k^{\ell^2} +2]  \big)$   & 98.06\%                    \\
    \hline
    \textbf{DVS Gesture} & &   \\
    Rate &\citep{shrestha2018slayer}   & $\mathcal{O}\big( H^{\ell}_{o} W^{\ell}_{o} C^{\ell}_{o} [ T + \tau C^{\ell-1}_{o}k^{\ell^2} ] \big)$ & 93.64\%                              \\
    Rate&\citep{kaiser2020synaptic}    & $\mathcal{O}\big( H^{\ell}_{o} W^{\ell}_{o} C^{\ell}_{o} [ T + \tau C^{\ell-1}_{o}k^{\ell^2} ] \big)$ & 95.54\%   \\
    Rate&\citep{fang2020exploiting}    & $\mathcal{O}\big( H^{\ell}_{o} W^{\ell}_{o} C^{\ell}_{o} [ T + \tau C^{\ell-1}_{o}k^{\ell^2} ] \big)$ & 96.09\%    \\
    Rate&\citep{he2020comparing}       & $\mathcal{O}\big( n^{\ell} [ \tau n^{\ell-1} + T ] \big)$    & 93.40\%    \\
    Rate&\citep{fang2021incorporating} & $\mathcal{O}\big( H^{\ell}_{o} W^{\ell}_{o} C^{\ell}_{o} [ T + \tau C^{\ell-1}_{o}k^{\ell^2} ] \big)$ & 97.57\%      \\
    Rate &\citep{cordone2021learning}   & $\mathcal{O} \big( T n_{a}^{t} C^{\ell}_{o} C^{\ell-1}_{o}  \big)$     & 92.01\%                   \\
    Temporal&DeNN ($q=0.5$)                        & $\mathcal{O}\big( C^{\ell}_{o} H^{\ell}_{o} W^{\ell}_{o} [\tau C^{\ell-1}_{o} k^{\ell^2} +2]  \big)$ &97.57\% \\   
    \hline
    \textbf{GSC} & & \\
    Temporal & \citep{hammouamri2023delays} & - & 95.35\% \\
    Temporal & \citep{bittar2024oscillations} & - & 97.05\% \\
    Temporal & \citep{deckers2024delays} & - & 95.69\% \\
    Rate     & \citep{wang2024convolutions} & - & 92.90\% \\
    Rate     & \citep{he2024msat} & - & 87.33\% \\
    Rate     & \citep{boeshertz2024mapping} & - &93.33\% \\
    Temporal & DeNN ($q=1$) & $ \mathcal{O}\big( n^{\ell}[ n^{\ell-1}+2+\nu ] \big) $ & 97.73\%
  \end{tabular} }
\end{table}
SpiNNaker power consumptions can be found in \citep{painkras2013spinnaker}. The energy consumption per instruction can be inferred as follows. The power of a SpiNNaker's chip is 1W peak. The idle total chip power is then made up of the idle chip power (0.36W) plus the SDRAM (0.170W), i.e., 0.53W. Of the remaining active power (1W - 0.53W = 0.47W), each link could use 0.063W and there are six, so 0.378W. Leaving, 0.47W - 0.378W = 0.092W, for core activity. There are 18 cores, so this is 0.0051W per core. Each clock cycle takes $5\cdot 10^{-9}$ seconds, so the energy consumption per clock cycle is then $2.56\cdot 10^{-11} J$.

From cores' technical documentation \cite{SpinnakerARM}, it can be found that each multiplication requires 2 clock cycles and each addition/subtraction 1 clock cycle. Using SpiNNaker own code \citep{partzsch2017exp}, based on fixed-point calculations \citep{partzsch2017exp}, each exponential function computation requires 95 clock cycles\footnote{Energy consumption results are extracted from a very interesting discussion with Andrew Rowley, researcher from the Human Brain Project in the School of Computer Science at the University of Manchester, and in charge of SpiNNaker developments.}.

Table \ref{energy_comp} presents the energy consumption on SpiNNaker neuromorphic supercomputer. 

Note that neuromorphic chips functioning and electronic architecture \citep{orchard2021loihi, akopyan2015truenorth, mayr2019spinnaker} are based on discrete event simulations. Many discrete time (synchronous) models, presented in Table \ref{energy_comp}, have the energy consumption advantage of not using exponential functions. However, their spike time is not exact, being up to the time step precision. In discrete event (asynchronous) models (like DeNN), spike times are exact but their computations come with the price of exponential function computations. However, many machine learning approaches require exponential functions. Currently, it is a problem taken seriously by electronics engineers. Energy consumption cost of an exponential function computation has recently been reduced to the insignificant cost of $3.63 \cdot 10^{-12}J$ for VLSI CMOS technology \citep{costa2023exponential}, which is used by neuromorphic computers.
\begin{table}[h!]
  \caption{Energy consumption of models. All symbols are the same as in Table \ref{complexities}.}
  \label{energy_comp}
  \centering
  \footnotesize
  \resizebox{\columnwidth}{!}{\begin{tabular}{ccc}
    \hline
    Model   & \vtop{ \hbox{\strut  Energy} \hbox{\strut Consumption} } & \vtop{ \hbox{\strut  On} \hbox{\strut SpiNNaker} }   \\
    \hline
    \textbf{MNIST} & &   \\
    \citep{zhang2020spike}    & $T\tau n^{\ell}n^{\ell-1}(2ADD+2MUL+IF)$ & $114T\tau\mu J$   \\
    \citep{kheradpisheh2020temporal}   & $T\tau n^{\ell}n^{\ell-1}(ADD+3IF)$   & $532\mu J$ \\
    DeNN ($q=1$)  &  $ n^{\ell} \big( \tau n^{\ell-1}(2EXP+3ADD)+2MUL+5ADD \big) + 2MUL $ & $73 \mu J$\\
    DeNN ($q=0.5$)  & $ n^{\ell} \big( \tau n^{\ell-1}(2EXP+3ADD)+2MUL+5ADD \big) + 2MUL $ & $\mathbf{40\mu J}$    \\
    \hline
    \textbf{CIFAR-10} & &  \\
    \citep{zhou2021temporal}    & $C^{\ell}_{o}H^{\ell}_{o}W^{\ell}_{o}\big( 3MUL + ADD[2\tau C^{\ell-1}_{o}k^{\ell^2}+1]+EXP[\tau k^{\ell^2}C^{\ell-1}_{o}]+IF \big) $  &  $620,190\mu J$\\
    DeNN ($q=1$)      & $C^{\ell}_{o}H^{\ell}_{o}W^{\ell}_{o} \big( \tau C^{\ell-1}_{o}k^{\ell^2}(2EXP+3ADD)+2MUL+5ADD \big) + 2MUL$    & $381,371\mu J$   \\
    DeNN ($q=0.5$)      & $  C^{\ell}_{o}H^{\ell}_{o}W^{\ell}_{o} \big( \tau C^{\ell-1}_{o}k^{\ell^2}(2EXP+3ADD)+2MUL+5ADD \big) + 2MUL$ & $\mathbf{232,161\mu J}$      \\
    \hline
    \textbf{N-MNIST} & &  \\
    \citep{zhu2022eventbackprop} & $TC^{\ell}_{o}H^{\ell}_{o}W^{\ell}_{o} \big( ADD( \tau c_{o}^{\ell-1}k^{\ell^2}+3) +2MUL+2EXP+IF \big)$ & $4301(0.9\tau+1)\mu J$ \\
    \citep{fang2021incorporating}& $TC^{\ell}_{o}H^{\ell}_{o}W^{\ell}_{o}\big( ADD( \tau C^{\ell-1}_{o}k^{\ell^2} + 3 ) + 2MUL + IF \big) $ &   $13,687\tau+375\mu J$          \ \\ 
    DeNN ($q=0.5$)                   & $ T \big[  C^{\ell}_{o}H^{\ell}_{o}W^{\ell}_{o} \big( \tau C^{\ell-1}_{o}k^{\ell^2}(2EXP+3ADD)+2MUL+5ADD \big) + 2MUL \big]$  & $\mathbf{11,616\mu J}$                   \\
    \hline
    \textbf{DVS Gesture} & &   \\
    \citep{fang2021incorporating} & $TC^{\ell}_{o}H^{\ell}_{o}W^{\ell}_{o}\big( ADD( \tau C^{\ell-1}_{o}k^{\ell^2} + 3 ) + 2MUL + IF \big) $ & $404,604\tau+11,267\mu J$ \\
    DeNN ($q=0.5$)            & $ T \big[  C^{\ell}_{o}H^{\ell}_{o}W^{\ell}_{o} \big( \tau C^{\ell-1}_{o}k^{\ell^2}(2EXP+3ADD)+2MUL+5ADD \big) + 2MUL \big]$ & $312,476\mu J$ \\   
    \hline
    \textbf{GSC} & & \\
    \citep{bittar2024oscillations} & $Tn^{\ell}\big( ADD[\tau(\frac{n^{\ell}}{2}+n^{\ell-1})+3] + 4MUL+COMP+ 2EXP + 2MUL + 2ADD \big)  $ & $39,4812(\tau + 0.21)\mu J$ \\
    \citep{deckers2024delays} & $ Tn^{\ell} \big( ADD[ \tau 2n^{\ell-1} +8 ] + 7MUL + COMP \big) $ & $25,005\mu J $ \\
    DeNN ($q=1$) & $ T \big[ n^{\ell} \big( \tau n^{\ell-1}(2EXP+3ADD)+MUL(\nu+2)+\nu EXP+5ADD \big) + 2MUL \big]$  & $\mathbf{20,715\mu J}$
  \end{tabular} }
\end{table}
\begin{table}[h!]
  \caption{Comparison of datasets, models and architectures in time and space dimensions.}
  \label{models_architectures}
  \resizebox{\columnwidth}{!}{\begin{tabular}{ccc|c|c|c|c|}
\cline{4-7}
\cellcolor[HTML]{FFFFFF}                            & \cellcolor[HTML]{FFFFFF}             & \cellcolor[HTML]{FFFFFF}                        & \multicolumn{4}{c|}{Model}                                                    \\ \cline{4-7}
                            &             &                        & ANN                      & IF   & LIF     & \multicolumn{1}{c|}{adLIF - DeNN} \\ 
\cmidrule{4-7}\morecmidrules\cline{3-7}                            &             & \multicolumn{1}{|c|}{\backslashbox[34mm]{Spatial Dim}{Time Dim} } & Null                     & Poor & Medium  & Strong                            \\ \cmidrule{3-7}\morecmidrules\cline{1-7}
\multicolumn{1}{|c|}{\multirow{3.5}{*}{\rotatebox[origin=c]{90}{\mytab{Architect.}}}}
& \multicolumn{1}{c||}{Fully Connected} & \multicolumn{1}{c||}{Null }                                           &  &      &         &                                   \\ \cline{2-7}
\multicolumn{1}{|c|}{}    
& \multicolumn{1}{c||}{}                & \multicolumn{1}{c||}{Poor}                                            &                          &      &         & \cellcolor[HTML]{656565} GSC                               \\ \cline{2-7}
\multicolumn{1}{|c|}{}  
& \multicolumn{1}{c||}{}                & \multicolumn{1}{c||}{Medium}                                          & MNIST                    &      &         &                                   \\ \cline{2-7}
\multicolumn{1}{|c}{}
& \multicolumn{1}{|c||}{Convolutional}   & \multicolumn{1}{c||}{Strong}                                         &  \cellcolor[HTML]{656565} CIFAR                    &      & Gesture &                                   \\ \hline
\end{tabular}}
\end{table}
\section{Distribution of neurons' spike times before standardization}
\label{sec:neurons_before_std}
\begin{figure}[h!]
    \centering
    \includegraphics[width=\linewidth]{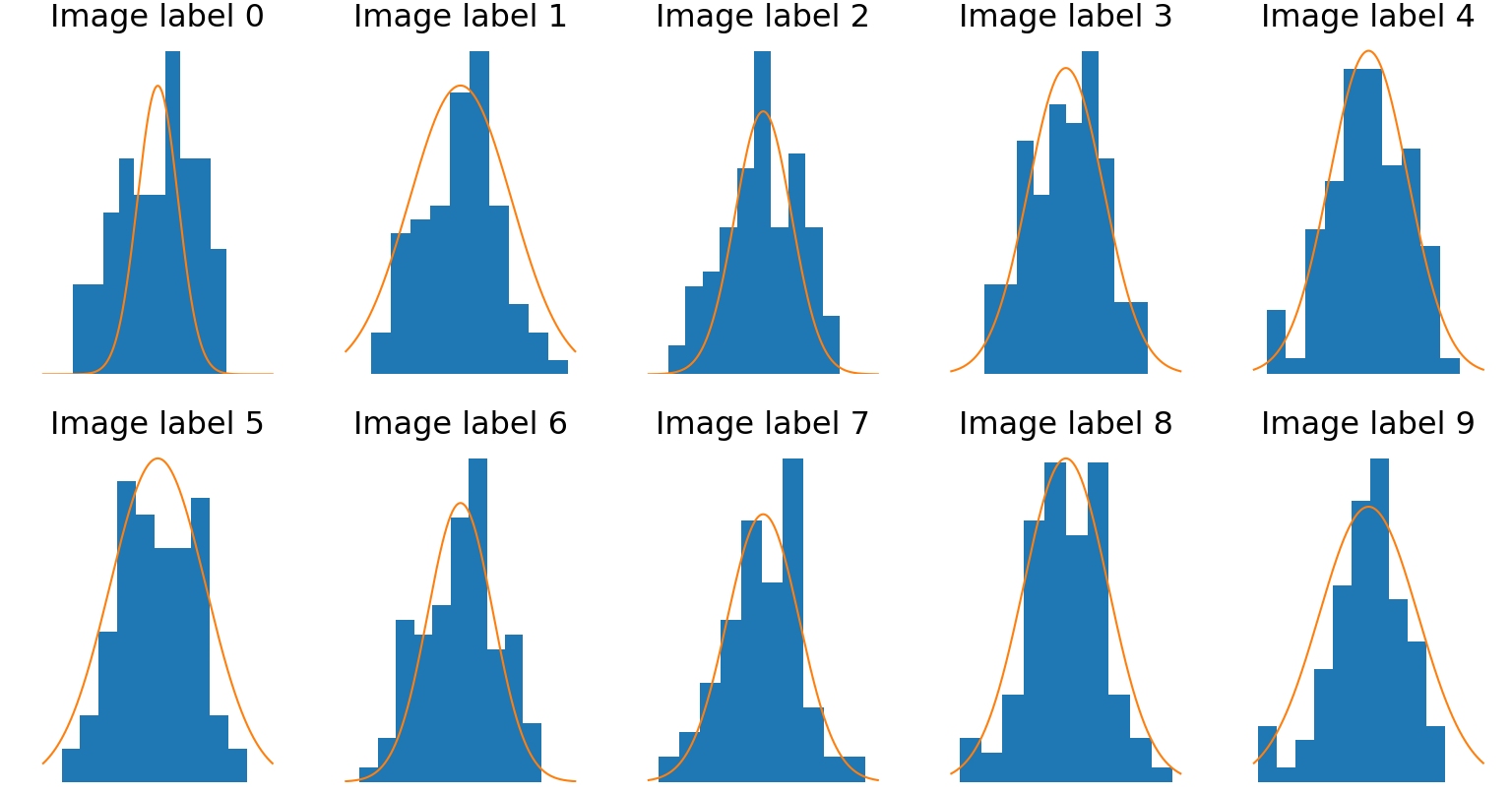}
    \caption{Distribution of neurons' spike times before standardization for the hidden layer of a network trained on the MNIST dataset, averaged over each class of images. Theoretical probability density function plot in solid orange line.}
    \label{fig:neurons_before_std}
\end{figure}

We show on figure \ref{fig:neurons_before_std} that the distributions of the spike times inside a layer, before the standardization process, is gaussian.
\section{Functioning of the network}
\label{suppsec:functioning}
\begin{figure}[h!]
    \centering
    \includegraphics[width=\linewidth]{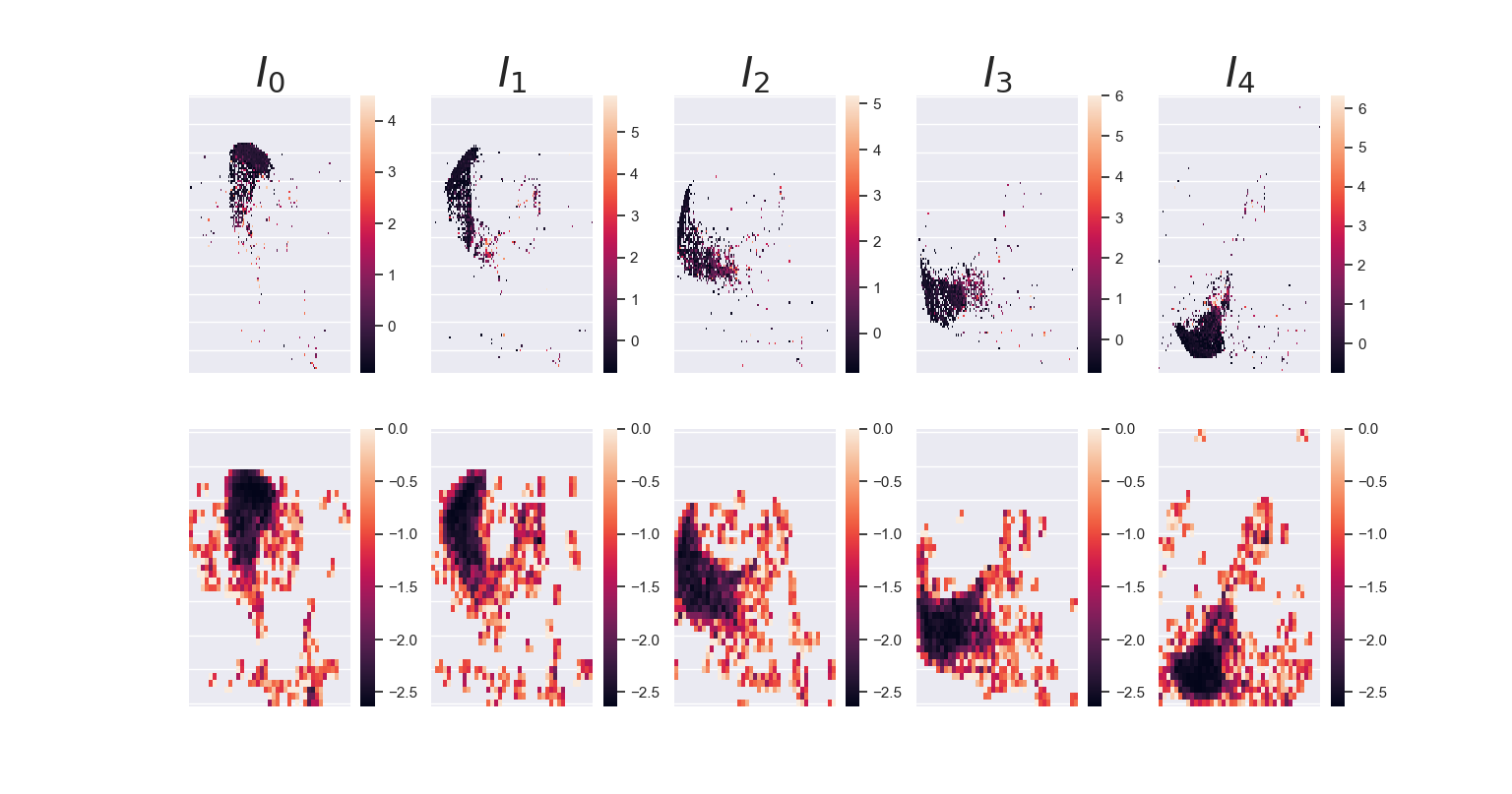}
    \caption{Top line: input images $I_{s}$. On bottom line: feature maps showing that the filter follows in time the areas that have the most activity. }
    \label{fig:fm_in_time}
\end{figure}

We show on Figure \ref{fig:fm_in_time} how a convolutional filter follows the activity in time.
\section{Synaptic impacts of spikes}
\label{sec:supp_matrices}
\begin{sidewaysfigure}[h!]
  \hspace*{-3cm}  
  \includegraphics[width=1.3\textwidth]{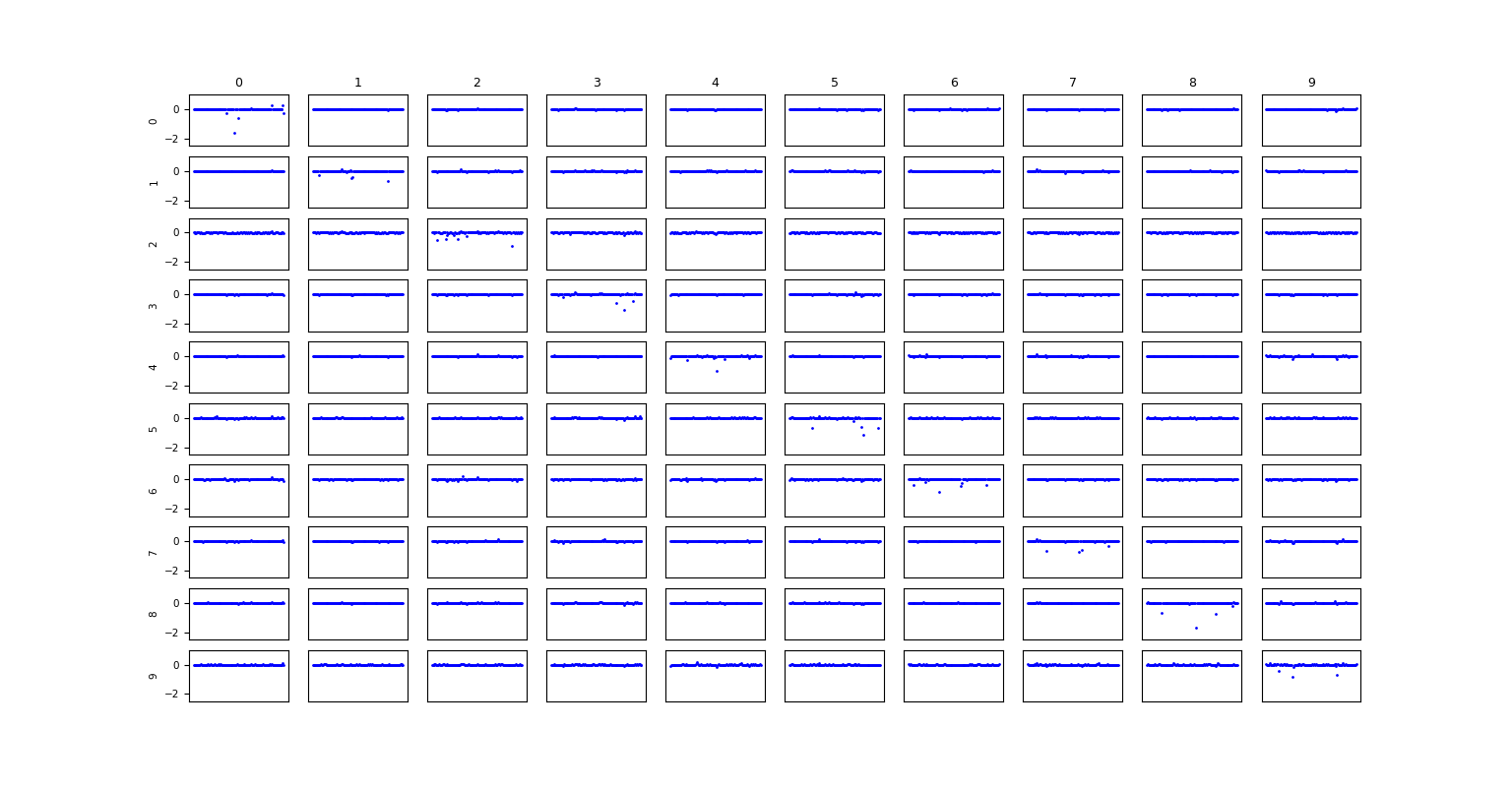}
  \caption{This grid of 10x10 graphs represents the 10 output neurons in columns for each image category in line, for the MNIST dataset. Each point on each graph represents the average synaptic impact of one intermediate neuron onto the output neuron, the image considered.}
  \label{matrix_mnist}
\end{sidewaysfigure}

\begin{sidewaysfigure}[h!]
  \hspace*{-3cm}  
  \includegraphics[width=1.3\textwidth]{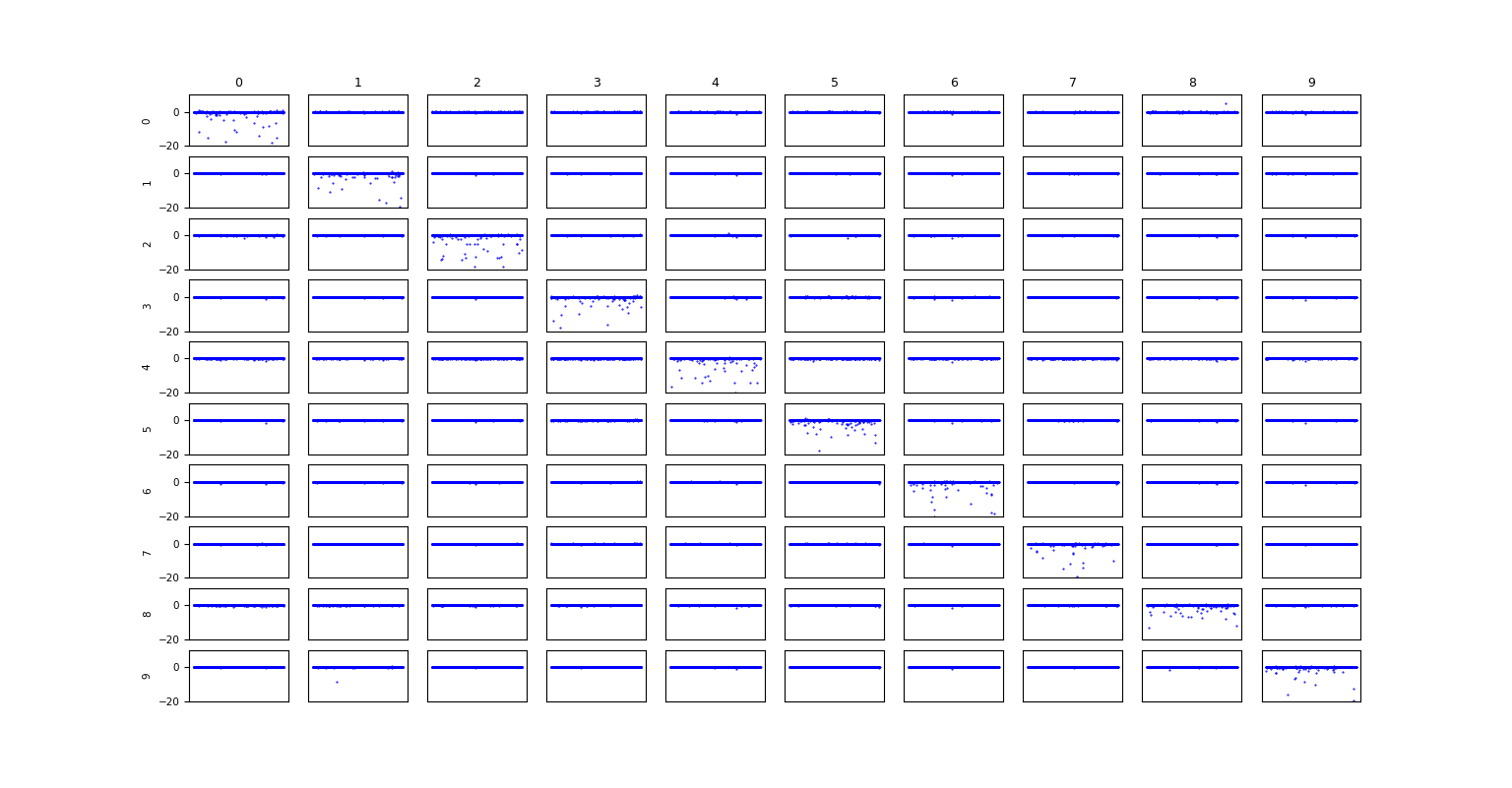}
  \caption{This grid of 10x10 graphs represents the 10 output neurons in columns for each image category in line, for the CIFAR10 dataset. Each point on each graph represents the average synaptic impact of one intermediate neuron onto the output neuron, the image considered.}
  \label{matrix_cifar}
\end{sidewaysfigure}

Figure \ref{matrix_mnist} presents the synaptic impact of each spike of the intermediary layer onto the output layer, for the fully connected DeNN for solving the MNIST task, and in Figure \ref{matrix_cifar} the same matrix is shown for the CIFAR-10 dataset.
\end{document}